\documentclass[default,iicol]{sn-jnl}

\usepackage{amsmath}
\usepackage{amssymb}
\usepackage{amsthm}
\usepackage{multirow}
\usepackage[T1]{fontenc}
\usepackage{xcolor}

\jyear{2021}%

\theoremstyle{thmstyleone}%
\theoremstyle{thmstyletwo}%

\theoremstyle{thmstylethree}%
\newtheorem{definition}{Definition}%

\begin{document}

\title[Deep transfer learning for image classification: a survey]{Deep transfer learning for image classification: a survey}

\author*[1]{\fnm{Jo} \sur{Plested}}\email{j.plested@unsw.edu.au}

\author[1]{\fnm{Musa} \sur{Phiri}\email{m.phiri@unsw.edu.au}}

\author[2,3,4]{\fnm{Tom} \sur{Gedeon}}\email{tom.gedeon@curtin.edu.au}

\affil[1]{\orgdiv{School of Engineering and Information Technology}, \orgname{University of New South Wales}, \country{Australia}}

\affil[2]{\orgdiv{AI Chair, Human-Centric Advancements}, \orgname{Curtin University},  \country{Australia}}
\affil[3]{\orgdiv{Honorary Professor of Computer Science} \orgname{Australian National University}}
\affil[4]{\orgdiv{International Research Professor in Informatics, John van Neumann School of Informatics} \orgname{Obuda University} \country{Hungary}}

\abstract{Deep neural networks such as convolutional neural networks (CNNs) and transformers have achieved many successes in image classification in recent years. It has been consistently demonstrated that best practice for image classification is when large deep models can be trained on abundant labeled data. However, there are many real world scenarios where the requirement for large amounts of training data to get the best performance cannot be met. In these scenarios, transfer learning can help improve performance. To date, there have been no surveys that comprehensively review deep transfer learning as it relates to image classification overall. We believe it is important for future progress in the field that current knowledge is collated and the overarching patterns analyzed and discussed. In this survey we formally define deep transfer learning and the problem it attempts to solve in relation to image classification. We survey the current state of the field and identify where recent progress has been made. We show where the gaps in current knowledge are and make suggestions for how to progress the field to fill in these knowledge gaps. We present new taxonomies of the solution and applications of transfer learning for image classification. These taxonomies make it easier to see overarching patterns of where transfer learning has been effective
and, where it has failed to fulfil its potential. This also allows us to suggest where the problems lie and how it could be used more effectively. We demonstrate that under this new taxonomy, many of the applications where transfer learning has been shown to be ineffective or even hinder performance are to be expected when taking into account the source and target datasets and the techniques used. In many of these cases, the key problem is that methods and hyperparameter settings designed for large and very similar target datasets are used for smaller and much less similar target datasets. We identify alternative choices that could lead to better outcomes.}


\maketitle

\section{Introduction}\label{Introduction}

It has been consistently shown that the success of deep neural networks for image classification relies upon an abundance of training data \cite{ngiam2018domain,mahajan2018exploring,kolesnikov2020big, cherti2023reproducible}.  However, there are many real world scenarios where this requirement  cannot be met. These include: 
\begin{enumerate}
\item Insufficient data because the data is very rare or there are issues with privacy etc. This scenario is common in the medical image domain \cite{kora2022transfer}.
\item It is prohibitively expensive to collect and/or label data. This is common for image tasks related to robotics. While large amounts of data can be created in simulation, when the model is transferred to real-world robots, it must be fine-tuned on limited data. This is because many robotics devices are expensive, and data needs to be collected in real time \cite{jaquier2025transfer}.  
\item The long tail distribution where a small number of classes are very frequent and thus easy to model, while many more are
rare and harder to model \cite{bengio2015sharing}. This is common with self-driving cars where usual driving conditions are well represented, but emergency type situations are rare \cite{wang2021advsim}.  
\end{enumerate}

There may be restraints on compute resources that limit training a
large model from random initialization with large amounts of data.
For example, environmental concerns \cite{strubell2019energy}. 

In all these scenarios transfer learning has the potential to significantly improve
performance. In this paradigm the model is pretrained on a related
dataset and task for which more data is available or, more recently, a more general task on the same dataset \cite{he2022masked, el2021large}. For this process to improve rather than harm performance the dataset must be related closely enough and best practice methods used.

With the exponentially increasing demand for
the application of modern deep neural networks to a wider array of real
world areas, work in transfer learning has increased at
a commensurate pace making it important to regularly take stock. In this paper we 
survey the current state of the field, where recent progress has been
made and where there are gaps in current knowledge. We also make suggestions
for how to progress the field to fill in these knowledge gaps. While
there are many surveys in related domains and specific sub areas,
to the best of our knowledge there are none that focus on deep transfer learning for image classification in general. We believe it is important
for future progress in the field that current knowledge is collated and the overarching patterns analyzed and discussed. 

We make the following contributions: 
\begin{enumerate}
\item Formally defining and categorizing deep transfer learning (DTL) and the problem it attempts to solve as it relates to image classification. Previous reviews have provided informal definitions of (DTL), and taxonomies \citep{tan2018survey,iman2023review,yu2022survey}. We refine these definitions and extend the taxonomies so they are more suited to modern (DTL) solutions and their application to image classification. 
\item Performing a thorough review and analysis of recent progress in the field. 
\item Presenting a continuous transition taxonomy highlighting the continuous nature of the source and target dataset relationships in transfer learning applications and their effect on the optimal solutions. This taxonomy helps us draw attention to the failure of transfer learning solutions to perform as expected in certain application areas. 
\item A new way of considering negative transfer so it can be eliminated in all scenarios. 
\item Giving a detailed summary of source and target datasets commonly used in the area to help the reader find the ideal source dataset to match their target task. 
\item Summarizing current knowledge in the area as well as pointing out
knowledge gaps and suggested directions for future research. 
\end{enumerate}
In Section \ref{sec:Related-work} we discuss the relationship between this survey and others in the area from general transfer learning to more closely related domains. In
Section \ref{sec:Overview} we introduce the problem domain and formalize
the difficulties with learning from small datasets that transfer learning
attempts to solve. This section includes terminology, definitions, and our new taxonomies that are referred to throughout this paper. Sections \ref{sec:inst_appr} to \ref{sec:best_prac} summarize current knowledge in relation to each of our five solution categories in our new taxonomy. Finally, Section \ref{sec:disc}
summarizes the overarching themes of these solutions, and Section \ref{sec:conc} summarizes our paper in full and recommends next steps for the field. 

\begin{table*}[th]

\caption{Transfer learning survey papers. Closely related papers are ones that sound most similar to ours based on the title but are limited in some way, broad are reviews of transfer learning in general not limited to image classification, and narrow are focused on a particular application of transfer learning. Under the narrow category, taxonomies are listed as NA for not applicable where their scope is limited, so any taxonomy would not be relevant to this work. For the same reason, the review is listed as limited even if it is a full review if it has a narrow scope, thus limited crossover with the scope of this paper. DTLIC refers to deep transfer learning for image classification  \label{survey_papers}}
\begin{tabular}{p{2.4cm}p{0.8cm}p{3cm}p{1.7cm}p{2cm}p{1.6cm}p{1cm}}
\hline 
Paper  & Year & Focus &  Deep learning focus & Image classification focus & Taxonomy & Full review \tabularnewline
 
\hline 
Closely related \tabularnewline
\cite{panda2024transfer} & 2024 & DTLIC & Yes & Yes & No & No \tabularnewline

\cite{ribani2019survey} & 2019 & DTLIC  & Yes & Yes & No & No \tabularnewline

Broad \tabularnewline

\cite{iman2023review}   & 2023 & DTL &  Yes & No & Yes & Yes \tabularnewline

\cite{yu2022survey} & 2022 & DTL & Yes & No & Yes & Yes \tabularnewline

\cite{tan2018survey} & 2018 & DTL & Yes & No & Yes & Yes \tabularnewline

Narrow \tabularnewline

\cite{hossen2025transfer} & 2025 & agriculture & Yes & No & NA & limited \tabularnewline

\cite{jaquier2025transfer} & 2025 & robotics & Yes & No & NA & limited \tabularnewline

\cite{kora2022transfer} & 2022 & medical images & Yes & limited & NA & limited \tabularnewline

\cite{li2020deep} & 2020 & facial expression recognition & Yes & limited & NA & limited \tabularnewline

\cite{minaee2020image} & 2020 & image segmentation & Yes & No & NA & limited \tabularnewline

\cite{mazurowski2019deep} & 2019 & radiology & Yes & limited & NA & limited \tabularnewline

\cite{ghosh2019understanding} & 2019 & image segmentation & Yes & No & NA & limited \tabularnewline

\hline 
\end{tabular}

\end{table*}

\section{Related work \label{sec:Related-work}}

Many reviews related to DTL have been published
in the past decade. The similarities and differences are summarized in Table \ref{survey_papers}. The existing reviews differ from ours in two important ways. The first group consists of more general reviews that provide a high level
overview of transfer learning and attempt to include all machine learning and task sub-fields. Reviews in this group are shown under the heading "Broad" in Table \ref{survey_papers}. The second group
is more specific, with reviews providing a comprehensive breakdown
of the progress on a particular, narrow, domain specific task. They are shown under the heading "Narrow" in Table \ref{survey_papers} and discussed further in the relevant parts of Sections \ref{sec:inst_appr} and \ref{sec:best_prac}. Our review is designed to sit in between these two groups and provides an overarching review of DTL as it relates to image classification specifically so that overarching patterns can be identified, analyzed and discussed. This allows us to make recommendations for best practice when using deep transfer learning for image classification (DTLIC), as well as identify knowledge
gaps and suggest directions for future research. 
A few surveys seem more closely related to ours, but they are not full reviews, only briefly introduce the area of DTLIC, and do not analyze recent advances in the field. These reviews are covered under the heading "Closely related" in Table \ref{survey_papers}.

\section{Overview \label{sec:Overview}}

Image classification tasks model relationships in complex hierarchical data. For example, there is no obvious relationship between raw RGB pixel values and whether the image is of a cat or a dog, and it would be impossible to hand-design a transformation that separates the two classes. For this reason, a large deep learning model is generally needed for image classification tasks, with millions or even billions of parameters arranged in layers that learn hierarchical combinations of features. The more complex the relationship between the input data and the prediction task, the more learnable parameters are needed to approximate the transformation function. 

\begin{figure}[t]
\centering

\includegraphics[scale=0.26, viewport=40bp 0bp 867bp 580bp, clip]{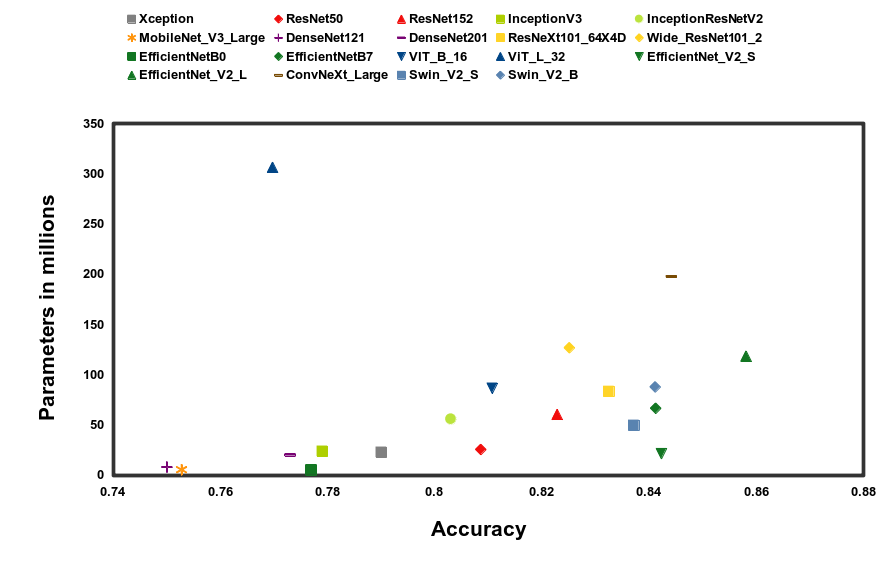}
\caption{Increase in performance on ImageNet 1K due to model size, measured
by number of parameters in millions. In general, larger models result in better performance. However, there are outliers that show a decrease in performance for the largest models. This is because the largest models overfit ImageNet1k. \label{fig:Increase-in-performance}}
\end{figure}

An example of complex hierarchical data is ImageNet 1k. Figure \ref{fig:Increase-in-performance} shows that, in general, the performance on ImageNet 1K \citep{imagenet_cvpr09} increases with the number of model parameters. However, some outliers show a decrease in performance as the model continues to increase in size. This is because the largest deep learning models overfit even the 1.3 million training examples in ImageNet1K \citep{ngiam2018domain,mahajan2018exploring,kolesnikov2020big,liu2022improved}. 

\subsection{Learning from small vs large datasets \label{subsec:Unreliable-Empirical-Risk}}

 When fitting large models to small, complex data, as is often needed with real-world data, it is essential to consider the classic \textbf{bias-variance trade-off}. We are attempting to minimize the error, which is the difference between the predictions of our model and the true output values. This difference is defined by a performance measure $P$ and the loss we are attempting to minimize $L$. We define $f^{*}$ as the true underlying function we are approximating. The error in our approximation originates from two places: 
\begin{enumerate}
    \item \textbf{Bias $\varepsilon_{bias}$}: the function that can be approximated is restricted by the choices we make regarding the model architecture. We define $\mathcal F$ as the set of functions defined by our model architecture and $f^{*}_\mathcal F$ as the function in $\mathcal F$ that minimizes our loss. This is unlikely to be the exact underlying function $f^{*}$, so our bias error becomes the difference between the optimal function in $\mathcal F$ and the true underlying function. We define this error as  $\varepsilon_{bias}$. Our bias error will be larger the more restrictive our model architecture is relative to the underlying function we are approximating, $f^{*}$. For example, if we try to approximate a higher order polynomial function like $x^2$ with a linear model $x$ we will have a high bias. 
    \item \textbf{Variance $\varepsilon_{var}$}: the difference between $f^{*}_F$ and the function that minimizes the loss from a particular set of training data. We define $L(f_n)$ as the empirical estimate of the loss given a particular dataset. So, our variance error becomes the difference between minimizing the empirical loss and the true loss for the optimal function in $\mathcal F$. We define this error as  $\varepsilon_{var}$. If we have a small number of examples in our training dataset and are attempting to model a complex function, our training data is unlikely to fully capture the function's complexities. 
\end{enumerate}

\noindent The total loss is then:
\begin{equation}
\label{eq:total_error}
    L = \varepsilon_{bias} + \varepsilon_{var} 
\end{equation}
\noindent being the sum of the bias loss coming from restrictions due to our choice in model architecture, and the variance loss coming from minimizing the empirical loss, based on the samples in our dataset, rather than the true underlying loss based on an infinite sample of all possible data and labels. 

It should be noted that the above errors depend entirely on the loss function we define. If our loss function does not accurately reflect the task we are trying to learn, this is a value alignment problem \cite{gabriel2020artificial}, which is beyond the scope of this discussion.

The bias error $\varepsilon_{bias}$ measures how closely
functions in $\mathcal{F}$ can approximate the optimal solution $f^{*}$. The variance error $\varepsilon_{var}$ measures the effect of
minimizing the empirical loss $L(f_{n})$ instead of the true underlying loss
loss $L(f^{*})$ \citep{bottou2008tradeoffs}. So finding a function
that is as close as possible to $f^{*}$ can be broken down into:
\begin{enumerate}
\item choosing a class of models that is more likely to contain the optimal
function
\item choosing training examples that better
approximate an infinite set of all possible data and labels.
\end{enumerate}

In general, $\varepsilon_{var}$ can be reduced by having a larger
number of examples \citep{wang2020generalizing}. Thus, when there are sufficient
and varied labeled training examples in our training dataset $D_{train}$ , the empirical loss
$L(f_{n})$ can provide a good approximation of the true underlying loss $L(f_{\mathcal{F}}^{*})$
the optimal $f$ in $\mathcal{F}$. When $n$ the number of training
examples is small the empirical loss may be further
from being a good approximation of the expected loss.
The resultant empirical loss minimizer overfits. For example, Ribeiro et al. created a curated dataset of images of wolves and huskies, ensuring that all wolves are on snow-covered backgrounds and all huskies are on snow-free backgrounds \citep{ribeiro2016should}. They showed that a neural network classifier trained on this dataset to predict whether an image shows a wolf or a husky focuses solely on the background, not on the animal itself. This is an extreme example of a model overfitting the idiosyncrasies of the dataset by minimizing the empirical loss, which is not a good approximation of the true underlying loss.  

To alleviate the problem of having an unreliable empirical loss
when $D_{train}$ is not sufficient, prior knowledge can be used to augment the data in $D_{train}$, constrain
the candidate functions $\mathcal{F}$, or constrain the parameters
of $f$ via initialization or regularization \citep{wang2020generalizing}.

In this work, we focus on transfer learning as a form of constraining the parameters of $f$ to address the unreliable empirical loss
problem. When weights pretrained on a large related dataset are used to initialize a deep neural network instead of random initialization, the loss landscape tends to be flatter and easier to navigate. This means that complex models with large numbers of trainable parameters are less likely to overfit to sharp local minima corresponding to idiosyncracies of the small target dataset \citep{liu2019towards}. Section \ref{sec:Areas-related-to} briefly discusses how deep transfer
learning relates to other techniques that use prior knowledge to solve
the small dataset problem.

\subsection{Deep transfer learning}

DTL is transfer learning applied to deep neural
networks. Much of the widely accepted terminology used to describe transfer learning comes from an early review \citep{pan2009survey}. The authors define transfer learning
as:
\begin{definition}
\label{def:transfer learning} Given a source domain $\mathcal{D\mathrm{_{S}}}$
and learning task $T_{S}$, a target domain $\mathcal{D\mathrm{_{T}}}$
and learning task $T_{T}$, transfer learning aims to help improve
the learning of the target predictive function $f_{T}\left(.\right)$
in $\mathcal{D\mathrm{_{T}}}$ using the knowledge in $\mathcal{D\mathrm{_{S}}}$
and $T_{S}$, where $\mathcal{D\mathrm{_{S}}}$$\neq\mathcal{D\mathrm{_{T}}}$, or $T_{S}\neq T_{T}$. 
\end{definition}

Tan et al. \citep{tan2018survey} define DTL as follows: 
Given a transfer learning task defined by $D_{S}, D_{T}, T_{S}, T_{T}$ and $f_{T}\left(.\right)$ . It is a deep transfer learning task where $f_{T}\left(.\right)$ is
a non-linear function that reflected a deep neural network.

For the purposes of this paper we refine this definition to be more precise: 

\begin{definition}
\label{def:deep transfer learning}
A transfer learning task is defined by $D_{S}, D_{T}, T_{T}$ and an optimal target function $f_{T}\left(.\right)$. It is a deep transfer learning task if 1) $f_{T}\left(.\right)$ is approximated by a deep neural network and 2) a deep neural network  $f_{S}\left(.\right)$ trained on the source dataset is used to improve the approximation of the target function $f_{T}\left(.\right)$.
\end{definition}
We define the source model as the deep neural network trained on the source dataset, the target model as the deep neural network trained on the target dataset, and the pretrained weights as the values of the weights after training on the source dataset is complete. Usually, DTL involves initializing some or all of the weights of the target model with the pretrained weights from the source model. If the source and target tasks are not identical then the final prediction layer weights must be randomly initialized rather than transfering them from the source model as the pretrained values will not be appropriate for the target task. 

\textbf{Fine-tuning } the weights is defined as continuing to train any transferred weights on the target task. 

\textbf{Freezing} the weights is defined as only training the newly initialized weights on the target task. 

\begin{figure*}[th]

\centering
\includegraphics[viewport=50bp 130bp 490bp 440bp, clip,scale=0.70]{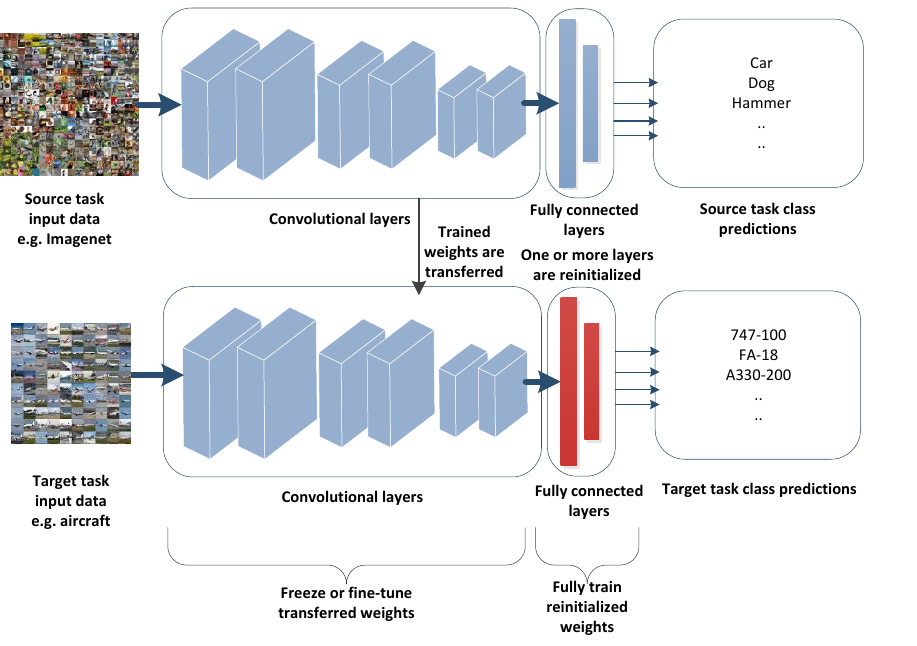}
\caption{Deep transfer learning\label{fig:Deep-transfer-learning}}
\end{figure*}

\noindent Any reinitialized weights must be trained. Figure \ref{fig:Deep-transfer-learning} shows the pretraining and fine-tuning pipeline of DTL. 

Combining the discussion from Section \ref{subsec:Unreliable-Empirical-Risk}
with definition \ref{def:deep transfer learning} using deep transfer
learning techniques to pretrain weights W can be thought of as regularizing
W. Initializing W with weights that have been well-trained on a large
source dataset rather than with very small random values makes it more difficult for the weights to move far from their pretrained values \citep{neyshabur2020being, liu2019towards}. In the most effective applications of DTL the source dataset is many orders of magnitude larger than the target dataset.
One example is pretraining on ImageNet 1K with 1.3 million training
images and transferring to medical imaging tasks which often only
have 100s of labeled examples. So even with the same learning rate
and number of epochs, the number of updates to the weights while training
on the target dataset will be orders of magnitude fewer than for pretraining.
When done well this prevents the model from learning weights that are based
on noise or idiosyncrasies in the small target dataset. When done poorly this can prevent the model from effectively learning features unique to the target dataset that could improve the model's performance. In the worst case scenario, the latter can result in negative transfer, which we describe in detail in Section \ref{sec:neg_tran}. 

\subsubsection{Analysis of the transfer learning optimization space}
Several feature space analysis methods were used in \citep{neyshabur2020being}. They found that models trained from pretrained weights make similar mistakes on the target domain, have similar features and are surprisingly close in $\ell_{2}$ distance in the parameter space. They are in
the same basins of the loss landscape. Models trained from random
initialization do not live in the same basin, make different mistakes,
have different features and are farther away in $\ell_{2}$ distance
in the parameter space.

A flatter and easier to navigate loss landscape for pretrained models
compared to their randomly initialized counterparts was also shown
in \citep{liu2019towards}. They showed improved Lipschitzness and
that this accelerates and stabilizes training substantially. Particularly
that the singular vectors of the weight gradient with large singular
values are shrunk in the weight matrices. Thus, the magnitude of the gradient
back-propagated through a pretrained layer is controlled and pretrained
weight matrices stabilize the magnitude of the gradient especially in
lower layers, leading to more stable training.

The above results mean that a pretrained model is likely to perform well and lead to more stable training when the source and target datasets are closely related. However, this stability means that it is more difficult to change the weights when they are not suited to the target task. This means that a much higher learning rate is usually needed \cite{dai2024adaptive}. 


\subsection{Learning from similar vs dissimilar datasets \label{sec:sim}}
In transfer learning it is important to match the source and target datasets as closely as possible. A more closely matched unlabeled source dataset will usually perform better than a less closely matched labeled dataset \citep{zoph2020rethinking,azizi2021big}. 

Dataset similarity refers to the quantifiable degree of shared characteristics between source and target datasets \citep{RN28, RN13}. The transferability of learned representations in deep neural networks is heavily influenced by dataset similarity, as greater similarity often enables more effective reuse of features. In contrast, large dissimilarities can lead to negative transfer, which is discussed in Section \ref{sec:neg_tran}. Similarity can be quantified through a similarity function $S: \mathcal{D} \times \mathcal{D} \rightarrow \mathbb{R}$, where $\mathcal{D}$ represents the space of all possible datasets, and  $S(d_1, d_2)$ yields a bounded, symmetric score indicating closeness between datasets \citep{RN104, RN103}.  

As noted by Stolte et al. \citep{RN26}, defining ``similarity'' is challenging when datasets differ in size, dimensionality, feature space, or label set. In some cases, source and target domain similarity can be effectively estimated through intuition and/or similarity between class labels \citep{mahajan2018exploring,ngiam2018domain,puigcerverscalable}. When this is not possible or we would like a more robust measure, there are many ways of measuring domain similarity.  Gupta et al. \citep{RN31} group similarity measurement techniques into four categories:
\begin{enumerate}
    \item \textbf{Correlation-based metrics}, such as the Pearson, Spearman, Kendall rank, and point-biserial correlation coefficients, quantify statistical associations between features. 
    \item \textbf{Distance-based metrics}, including Euclidean, Manhattan, Mahalanobis, Canberra, Levenshtein, Wasserstein, Fréchet distance (FD), Earth Mover's Distance (EMD), and Hamming distances, assess geometric differences in feature space.
    \item \textbf{Probability-based measures}, such as mutual information, conditional probability, and divergence metrics including Kullback--Leibler, Jensen--Shannon, Bhattacharyya, Maximum Mean Discrepancy (MMD), Hellinger, and total variation distance, compare underlying data distributions.
    \item \textbf{Set-based measures}, such as the Jaccard Index, Dice Coefficient, Cosine Similarity, and Overlap Coefficient, evaluate overlap in categorical or discrete representations.
\end{enumerate}

 These measures provide complementary perspectives, and the choice of method depends on the data modality, domain, and intended transfer task.  

In practice, the suitability of a similarity measure depends strongly on dataset characteristics and the nature of the domain gap.

MMD has proven especially effective for aligning high-dimensional visual domains with distribution shifts but similar underlying semantics, successfully enabling knowledge transfer between medical imaging modalities such as CT and MRI scans \citep{RN107,RN110}. For scenarios requiring optimal transport solutions, such as few-shot settings where cluttered backgrounds, large intra-class variations and the loss of local structural information make global embeddings unreliable, EMD excels by comparing structured representations through optimal matching \citep{RN109}. This approach has demonstrated strong performance on challenging datasets such as miniImageNet and tieredImageNet \citep{RN109}. Wasserstein distance proves particularly valuable for applications where source and target domains exhibit minimal distributional overlap, such as transferring knowledge from simulation data to real sensor measurements in fault diagnosis systems \citep{RN108}. When computational efficiency is paramount, Bhattacharyya distance provides effective similarity estimation without requiring expensive fine-tuning procedures, making it ideal for rapid model selection in computer vision applications \citep{RN182}.

These examples highlight how data modality, degree of semantic alignment, and size of the domain gap inform the optimal choice of similarity measure for deep transfer learning.

\subsection{Example showing the effects of target dataset size and similarity on deep transfer learning}

To motivate our discussions on the importance of considering the size of the target dataset and its similarity to the source dataset in determining the optimal transfer learning solution for a given target task we present a case study where deep transfer learning fine-tuning hyperparameters designed for moderately sized target datasets that are extermely similar to the target dataset perform poorly on smaller, dissimilar target datasets.  

\paragraph{Big Transfer - transferring from larger source datasets to small, similar target datasets}
Big Transfer (BiT) was defined as pretraining on source datasets
larger than ImageNet 1K in \citep{kolesnikov2020big}. The authors devise general fine-tuning hyperparameter rules that are only dependent on the size of the target dataset. They show that these rules result in close to optimal performance for a range of datasets that are either 
extremely similar to the source datasets or, in the case of Flowers, known to respond well to protocols designed for extremely similar target datasets. They also show that decaying the learning rate faster works better for moderately sized compared to large target datasets.
We hypothesise that these best practice recommendations will be far from optimal for less closely related target datasets, as has been shown with many other methods and recommendations
\citep{xuhong2018explicit,li2019delta,wan2019towards,plested2021rethinking,li2020rethinking,chen2019catastrophic,kou2020stochastic}.

\subsubsection{Experiment extending Big Transfer to less similar datasets. }

An extension to the Big Transfer experiments was performed to show how training with hyperparameters recommended for small, target datasets similar to the source dataset performs on two less similar target datasets. We chose two datasets that are known to be difficult to transfer to and conceptually different from our source dataset, ImageNet21K. These are Stanford Cars (Cars) \cite{kraus2017automated} and FGVC Aircraft (Aircraft) \cite{maji13fine-grained}. We first measure the similarity of all target datasets to ImageNet21K using three measures, EMD \cite{RN109}, Wasserstein distance \cite{RN108}, and FD described in Section \ref{sec:sim}. The results of these measures are shown in Table \ref{sim_results} and show that, as expected, our two new datasets have much lower similarity measures than the original datasets.

\begin{table*}[th]
\caption{Dataset similarity and distance measures for five target datasets compared to ImageNet1k as the source dataset. The similarity measures are normalized so that 1.0 is identical and 0 is no similarity and distance scores are unnormalized with higher distances meaning the datasets are less similar. The measures show that the two datasets that are conceptually similar to ImageNet1K, being Cifar10 and Cifar100 have higher similarity scores and lower distance measures than those that are less conceptually similar, Cars and Aircraft, and known to be harder to transfer to from ImageNet1K. \label{sim_results}}
\begin{tabular}{p{1.8cm}p{1.8cm}p{1.8cm}p{2cm}p{2cm}p{1.8cm}p{1.8cm}}
\hline 
Target Datasets	& EMD \hspace{1cm} (distance) &	EMD \hspace{1cm} (similarity) &	Wasserstein (distance) &	Wasserstein (similarity) &	FD \hspace{1cm} (distance) &	FD  \hspace{1cm} (similarity) \tabularnewline
 
\hline
Cifar10 &	0.9755 & 0.6140 &	893.2167	 & 0.6398 &	755.6660 &	0.6749 \tabularnewline
Cifar100	& 0.8783	& 0.6446	 & 760.7420	 & 0.6836 &	511.1466 &	0.7665 \tabularnewline

Oxford pets &	1.0362 &	0.5956 &	1012.8434 &	0.6026 &	1238.4232 &	0.5250 \tabularnewline
Cars &	1.1286 &	0.5688 &	1152.1820	& 0.5621 &	1323.2090 &	0.5024 \tabularnewline
Aircraft &	1.2809 &	0.5270 &	1278.8186 &	0.5276 &	1415.3668 &	0.4788 \tabularnewline

\hline 
\end{tabular}

\end{table*}

 We then trial different fine-tuning hyperparameters than those recommended for more similar target datasets. The results are shown in Table \ref{BiT}. It can be seen that the recommended learning rate is effective for the most similar datasets, being Cifar10, Cifar100, and Oxford Pets. However, it is far from optimal for the less similar datasets, being Aircraft and Cars. While the decay schedule does not make a significant difference to final mean accuracy, it does produce more robust results with far smaller standard deviations in both cases. So, when looking for one best final prediction model, the decay schedule should be considered, whereas for an ensemble of models, it may be less important. 

For the best and most robust results, the similarity of the source and target datasets, in addition to the size of the target dataset, should be taken into account when setting fine-tuning hyperparameters. It should be noted that the effect of using a learning rate suited to more similar target datasets on less similar target datasets is much more pronounced than vice versa. In general, small, less similar target datasets are more sensitive to transfer learning protocols. 

\begin{table*}[th]

\caption{Extension of Big Transfer (BiT) to less similar datasets \cite{kolesnikov2020big} \label{BiT}. All results are obtained using a ResNet-50x1 pretrained on ImageNet-21k (BiT-M). Cifar10 \cite{krizhevsky2009learning}, Cifar100 \cite{krizhevsky2009learning} and Oxford Pets \cite{parkhi2012cats} are examples of datasets from the original paper that are very similar to ImageNet-21k, while Cars \cite{KrauseStarkDengFei-Fei_3DRR2013} and Aircraft  \cite{maji13fine-grained} are less similar. Recommended values are those values recommended for learning rate and learning rate decay from the paper. Less similar learning rate (lr) and less similar decay are the best lr (lr=0.02) and lr decay (lr decay steps=3, rate=0.97) for both less similar datasets overall. Less similar combined is the combination of the best lr and lr decay for less similar datasets. Best result is the best result with any hyperparameters for the given dataset individually, taking into account both mean accuracy and standard deviation}
\begin{tabular}{p{1.2cm}p{2cm}p{2cm}p{2cm}p{2cm}p{2cm}p{1.7cm}}
\hline 
Dataset  & recommended values &  less similar lr & less similar decay & less similar combined  & best result & best hyperparameters \tabularnewline
 
\hline
Cifar10 & 97.626±0.108 & 97.402±0.069 & 97.014±0.019 & 97.086±0.108  &  97.626±0.108 & lr=0.003,	lr decay steps=100, rate=0.1\tabularnewline

Cifar100 & 86.93±0.210 & 84.977±0.224  & 86.484±0.149 & 85.238±0.263 & 86.93±0.210  & lr=0.003,	lr decay steps=100, rate=0.1\tabularnewline

Oxford pets & 92.856±0.095 & 91.538±0.806 & 92.914±0.138 & 92.026±0.082 & 92.856±0.095 & lr=0.003,	lr decay steps=100, rate=0.1\tabularnewline

Cars &  81.812±0.151
 & 87.602±0.692
 & 81.825±0.216 & 87.512±0.335 & 87.958±0.239 & lr=0.025,	lr decay steps=5, rate=0.96
\tabularnewline
Aircraft & 82.414±0.177
 & 87.256±4.750 & 81.448±0.256 & 86.95±0.308 & 86.95±0.308 & lr=0.02, lr decay steps=3, rate=0.97
\tabularnewline

\hline 
\end{tabular}

\end{table*}

\subsection{Negative Transfer \label{sec:neg_tran}}

If the source dataset is not well related to the target dataset the target model can be negatively impacted
by pretraining. This situation is called negative transfer.

Wang et al. elaborate on factors
that affect negative transfer \citep{wang2019characterizing,wang2019transferable}:

\begin{itemize}
\item Similarity between the source and target domains. Transfer learning
assumes that there is some similarity between joint distributions
in source domain $P_{S}(X,Y)$ and target domain $P_{T}(X,Y)$. The
higher the divergence between these values the less information there is
in the source domain that can be exploited to improve performance
in the target domain. In the extreme case if there is no similarity,
it is not possible for transfer learning to improve performance.
\item The size and quality of the source
and target datasets. The excess empirical loss bounds of the pretrained model are shown to be dependent on the size of the target dataset in \citep{liu2022improved}. Intuitively, if labeled target data is abundant
enough, a model trained on this data only may perform well. In this example,
transfer learning methods are more likely to impair the target learning
performance. Conversely, if there is no labeled target data, as in the domain adaptation task, a bad transfer learning method would likely perform better than a random guess, which means negative transfer would not happen.
\end{itemize}
The importance of similarity between the source and target datasets and the size of the target dataset is shown when transferring between ImageNet 1k and the COCO object detection dataset (200,000 images and 1.5 million labeled object instances). Experiments in \cite{he2018rethinking, zoph2020rethinking} show that using weights pretrained on ImageNet 1K results in negative transfer in many experimental settings. Further results show that doing self-supervised pretraining on the target dataset only, never results in negative transfer \cite{zoph2020rethinking}. These results are expected as the source and target datasets are quite dissimilar as they are entirely different tasks, and COCO has slightly more labeled examples than ImageNet 1K.   

With deep neural networks, once the weights have been pretrained to
respond to particular features in a large source dataset it will be difficult to change these weights far from their pretrained
values \citep{neyshabur2020being}. This is particularly so if the
target dataset is orders of magnitude smaller as is often the case.
This premise allows transfer learning to improve performance and also
allows for negative transfer. If the weights transferred are pretrained
to respond to unsuitable features then this training will usually not be fully
reversed during the fine-tuning phase and the model could be more
likely to overfit to these inappropriate features.

Most cases of poor and negative performance for transfer learning occur when all the following elements are present: 
\begin{enumerate}
    \item The target dataset is large enough to be trained reasonably well from random initialization \citep{he2018rethinking}.
    \item   The target and source datasets are sufficiently dissimilar that weights pretrained on the source dataset are unlikely to be a good fit for the target dataset \citep{he2018rethinking}.
    \item The transfer learning solutions used do not allow the weights to change far enough from their pretrained values. This point is further discussed in Section \ref{sec:best_prac} on best practice.  
\end{enumerate} 

In image classification models, features learned through lower layers
are more general, and those learned in higher layers are more task
specific \citep{yosinski2014transferable, leesurgical}. It is likely that if less
layers are transferred negative transfer should be less prevalent,
with training all layers from random initialization being the extreme
end of this. However, there has been limited prior work to test this.

\subsection{Categorization of Transfer Learning}

\begin{figure*}
\centering
\includegraphics[viewport=30bp 530bp 600bp 830bp, clip, scale=0.60]{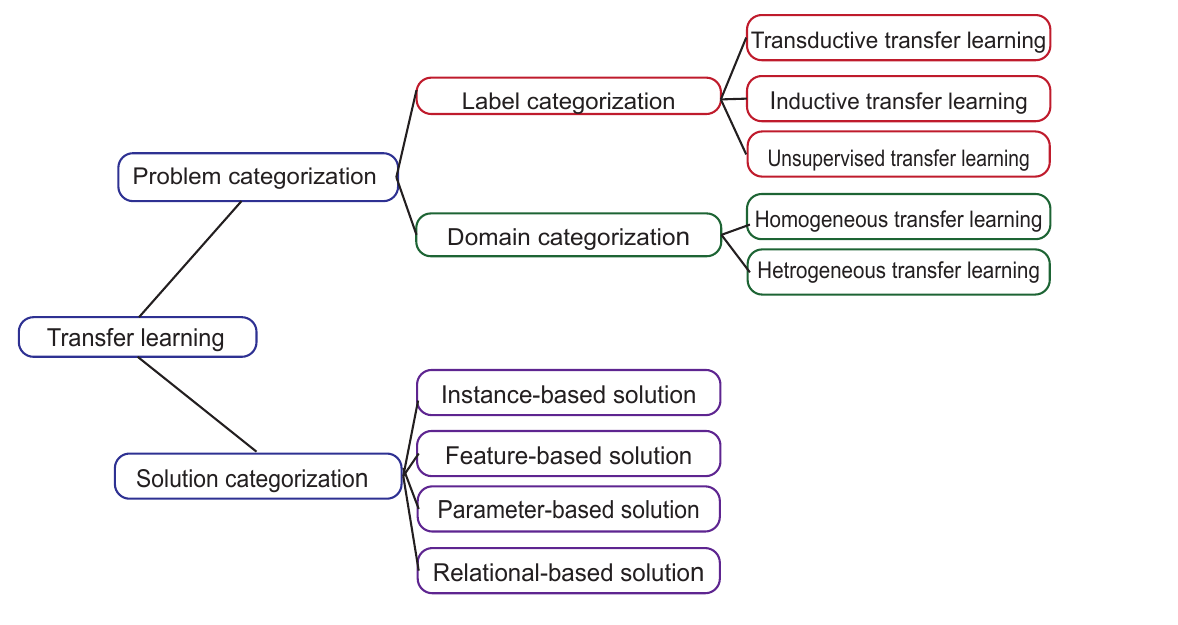}
\caption{General transfer learning taxonomy \citep{zhuang2020comprehensive} \label{fig:taxonomy-general}}
\end{figure*}

General transfer learning has traditionally been categorized in several ways see Figure \ref{fig:taxonomy-general} for an overview. These categorizations include: 
\begin{enumerate}
    \item Label based categorization which looks at the availability of labels and divides transfer learning into the following categories: 
        \begin{itemize}
            \item Inductive transfer learning - the source and target tasks are different and labels are available for at least the target dataset. Labels may or may not be available for the source domain.   
            \item Transductive transfer learning - the source and target tasks are the same but the domains are different. Labels are available for the source dataset only. Referred to as domain adaptation.
            \item Unsupervised transfer learning - labels are not available for either dataset.
    \end{itemize}

    \noindent For a full exploration of this categorization see  \citep{pan2009survey}. 
    \item Consistency between source and target data \citep{weiss2016survey}. 
    \begin{itemize}
        \item When $\mathcal{X\mathrm{_{S}}} = \mathcal{X\mathrm{_{T}}}$
and $y_{S} = y_{T}$ this is referred to as homogeneous transfer learning.
        \item When $\mathcal{X\mathrm{_{S}}}$$\neq\mathcal{X\mathrm{_{T}}}$
or $y_{S}\neq y_{T}$ this is heterogeneous transfer learning.
    \end{itemize}
    Traditionally, homogeneous transfer learning is said to have the same feature and input space. However, when extending this concept to DTL care must be taken as the feature space is learned from the data. It is common in DTL for computer vision to have $\mathcal{X\mathrm{_{S}}}$$=\mathcal{X\mathrm{_{T}}}$. This means usually either the input dimensions and ranges are equal or they are preprocessed to be. However usually $\mathcal{P(X\mathrm{_{S}})}$$\neq\mathcal{P(X\mathrm{_{T}})}$ and $y_{S}\neq y_{T}$ meaning the distributions of input and target values are different; therefore new features must be learned from the target dataset. 

    \item Solution categorization \citep{pan2009survey, zhuang2020comprehensive} divides transfer learning into categories based on approaches to solving the task:
    \begin{itemize}
        \item Instance-based approaches - mostly based on weighting instances in the source dataset when training on the target task.
        \item Feature-based approaches - transform the source features to create a new feature representation for the target task, or learn a shared feature space.
        \item Parameter-based approaches - transfer knowledge from the source to the target dataset via the model/parameters.
        \item Relational-based approaches - deal with the relational domain, generally focusing on transferring the relationship via logical rules. 
    \end{itemize}
    In DTL by definition, the features are learned and in some form the pretrained model is always used. As a result, it can be argued that all deep learning methods fall under parameter-based approaches as per \citep{zhuang2020comprehensive}. The other three approaches have been used successfully in conjunction with parameter-based approaches \citep{zhu2021transfer, ngiam2018domain,mahajan2018exploring}. When considering feature-based approaches it is important to keep in mind that deep learning features are learned, in contrast to precalculated features used by traditional machine learning approaches. This means that feature-based approaches for DTL generally involve learning a shared feature space or methods that restrict or encourage the source and target feature spaces to be similar. A recent successful example is the CLIP algorithm for text to image generation \citep{radford2021learning}. This algorithm encourages the text embedding space to be as close as possible to the image embedding space.  It uses a contrastive loss that maximizes the dot product of the two embeddings if the caption belongs to the image and minimizes it if it belongs to a different image. 
\end{enumerate}

\subsection{Categorization of Deep Transfer Learning for Image Classification \label{sec:categ}}
As shown above, the traditional categorizations of transfer learning are not always well suited when applied to DTL, particularly in relation to image classification.  We therefore propose a new categorization.

From Section \ref{subsec:Unreliable-Empirical-Risk} we have $f_{\mathcal{F}}^{*}$ the optimal function in $\mathcal{F}$ the family of functions defined by our model architecture. By Definition \ref{def:transfer learning} we have $\mathcal{D\mathrm{_{S}}}$$\neq\mathcal{D\mathrm{_{T}}}$, or $T_{S}\neq T_{T}$.  This implies that the optimal function for the source task is not equal to the optimal function for the target task $f_{S\mathcal{F}}^{*} \neq f_{T\mathcal{F}}^{*}$  and means that the optimal weights of the model must be learned based on the target dataset. Usually in the transfer learning setting, the target dataset is orders of magnitude smaller than the source dataset. From equation \ref{eq:total_error} we are trying to minimize the sum of the approximation error $\varepsilon_{app}$ and the estimation error $\varepsilon_{est}$. If the relationship between the input data and labels is complex a model with a large number of parameters (usually in the millions or billions) is needed to ensure that the optimal function in $\mathcal{F}$ is as close as possible to the optimal function in general and reduce $\varepsilon_{app}$. The larger the model the more capacity it has to overfit and the larger $\varepsilon_{est}$ can become.  This means deep learning for complex tasks like image classification involves selecting a model that is large enough to learn the relationships between the input and label data and regularizing it to ensure the effect of $\varepsilon_{est}$ is reduced as far as possible. Pretraining a large deep learning model is a form of regularization as it creates a flatter optimization landscape with smaller gradients \citep{liu2019towards}. This allows for small updates to the weights based on the target dataset while discouraging large updates. The effect can be magnified by adding a regularization term that explicitly decays the weights towards their pretrained values \citep{xuhong2018explicit}.

Consequently, with DTLIC in particular, we end up with a balance between:
\begin{enumerate}
    \item allowing the pretrained weights to move towards the weights of the optimal function of the target task $f_{T\mathcal{F}}^{*}$.
    \item preventing the new weights from overfitting the empirical loss approximation $R(f_{n})$. 
\end{enumerate}

\begin{figure*}
\centering
\includegraphics[viewport=40bp 285bp 700bp 590bp, clip, scale=0.60]{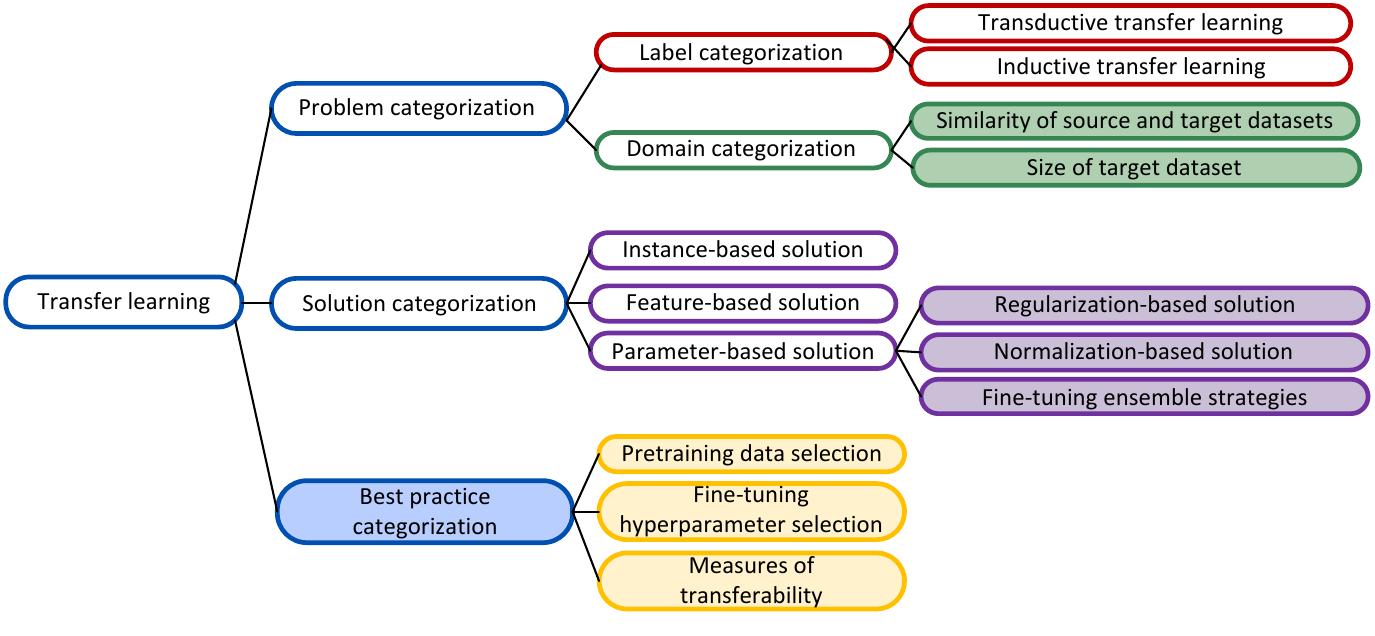}
\caption{Deep learning for image classification taxonomy. Shaded cells are additions to existing transfer learning and deep transfer learning taxonomies.\label{fig:taxonomy-new}}
\end{figure*}

Our new categorization of DTLIC, as shown in Figure \ref{fig:taxonomy-new}, both truncates and expands on the traditional transfer learning categories.

\subsubsection{Domain Categorization \label{sec:domain_cat}} 
As we are concerned with the final task of image classification it is assumed that we have labeled examples in: 
\begin{itemize}
    \item the source domain but not the target domain - domain adaptation  
    \item the target domain but not the source domain - self-supervised pretraining
    \item both domains
\end{itemize}
Given this assumption, we removed the unsupervised transfer learning setting from the label based categorization.

We define two new categories.

\textbf{Similarity}:  In DTL the features are learned from the data. The categorization of homogenous versus heterogeneous thus becomes a solution categorization of whether to continue training the pretrained weights or not during the fine-tuning process so we remove it. We argue that the optimal decision in the solution space still depends on the consistency between source and target data - including both the input data and task. The more $f_{T\mathcal{F}}^{*}$ differs from $f_{S\mathcal{F}}^{*}$ the less likely that $f_{S\mathcal{F}}^{*}$ will be an optimal substitute for $f_{T\mathcal{F}}^{*}$. To represent this in our taxonomy, we replace homogeneous versus heterogeneous with similar versus dissimilar. 

\textbf{Size:}  The balance described above of allowing the weights to move but not overfit points to another important categorization in DTL for image classification -  the size of the dataset. The larger the model compared to the size of the target dataset the more likely it is to overfit if the weights are allowed to move further from their pretrained values. To represent this element we added large versus small dataset under domain categorization. 

\begin{figure*}[th]
\centering
\includegraphics[viewport=0bp 48bp 605bp 550bp,clip,scale=0.45]{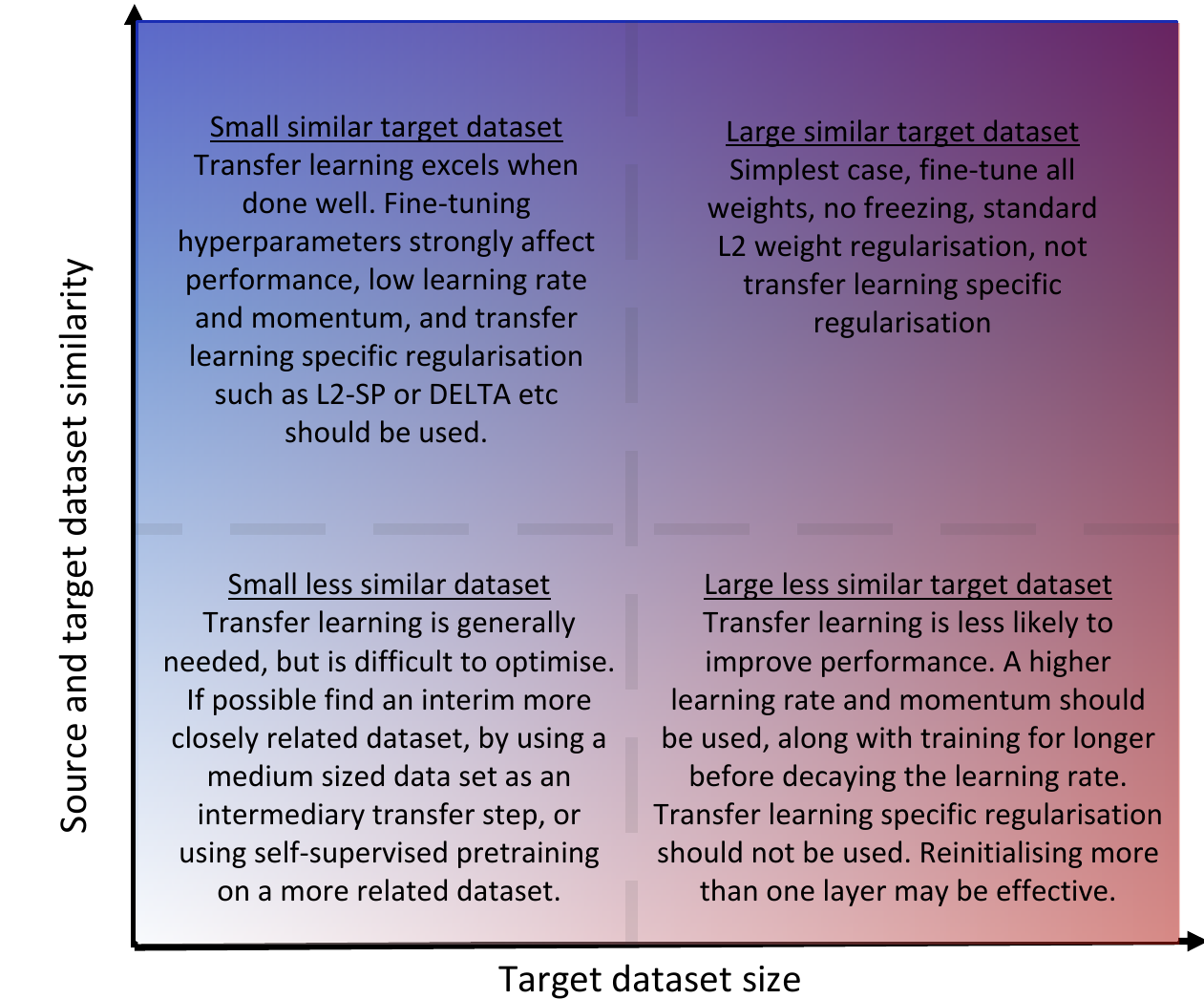}\newline
\caption[Transfer learning taxonomy]{Transfer learning taxonomy. Blue is a measure of source and target dataset similarity, where the most similar would be transferring from one set of ImageNet classes to another such as in \cite{yosinski2014transferable,plested2019analysis} or self-supervised pretraining on the target dataset (self-training). Red measures the target dataset size,  with small being around 10,000 labeled examples
and large being a million plus labeled examples. The lines between the quadrants are gray and dotted and the colors gradually merge to indicate that there is not a clear cutoff point from one quadrant to the next, rather as the target dataset increases
in size between the range of around 50,000 to 200,000 more strategies for large target
datasets should be employed. The number of trainable parameters in the model architecture should be considered when deciding whether to use strategies for small or large datasets. Larger models generally require more training data to avoid overfitting the target task, as highlighted in Figure \ref{fig:Increase-in-performance} \label{tax}.}
\end{figure*}

The two new categorizations above are not discrete so we also present Figure \ref{tax} representing them on a continuous scale. The insights shown in this diagram are discussed further in Section \ref{sec:best_prac}.

\subsubsection{Solution categorization}

In the solution categorization space, we remove relational-based approaches. On the rare occasions when relational-based tasks are seen in the image classification task domain, their solutions generally involve the transfer of learned parameters, for example, graph neural network approaches, not logical rules. 

Some previous DTL taxonomies \citep{tan2018survey, iman2023review, yu2022survey} include a generative adversarial training (GAT) categorization. This refers to a wide variety of techniques that use adversarial training to improve results in transfer learning tasks \citep{tzeng2017adversarial, shrivastava2017learning, liu2016coupled, attia2020realistic}. We argue that the significant uses of GAT in transfer learning are better described by alternative solution categories or by areas only peripherally related to DTLIC. 

The use of GAT for transfer learning often falls into other solution categories depending on its use. For example, aligning the source and target feature spaces \citep{tzeng2017adversarial, liu2016coupled, gamrian2019transfer} and better matching of synthetic data with real data \citep{shrivastava2017learning, taigman2017unsupervised} are feature-based approaches and are discussed in Section \ref{sec:fea_appr}

GATs can also be used in ways peripheral to transfer learning, such as for data augmentation. For example, generating new conditional samples from the source domain that are as similar as possible to the target domain \citep{attia2020realistic}, or improving the realism of synthetic samples generated another way \citep{shrivastava2017learning} are examples of data augmentation. While they can be effective when applied to transfer learning, these techniques are beyond the scope of this survey. See \citep{biswas2023generative} for a detailed overview of the use of data augmentation using GATs for image classification.

As many DTLIC approaches can be classed as parameter-based approaches, we further categorize them as follows: 

 \textbf{Regularization based approaches}: assume that the optimal solution is not too far from the pretrained solution and so the model weights or features should be prevented from moving too far during fine-tuning. 
    
 \textbf{Normalization based approaches:}  seek to better align fine-tuning in the target domain with the source domain by making adjustments to the standard batch normalization
    or other forms of normalization that are used
    between layers in modern deep learning models.
    
\textbf{Fine-tuning ensemble strategies} combine multiple fine-tuning or weight freezing strategies to optimize the final fine-tuned model. 

Note that some previous DTL surveys have included solution categories for fine-tuning and freezing weights \citep{iman2023review, yu2022survey}. These are fine-tuning hyperparameter decisions and fall under our new categorization of best practice introduced below.

\subsubsection{Best practice categorization}

Finally, we add a new root categorization of best practice. To this, we add the following subcategories: 
\begin{itemize}
    \item Pretraining data selection
    \item Fine-tuning hyperparameter selection
    \item Measures of transferability.
\end{itemize}

\subsection{Areas related to deep transfer learning for image classification
\label{sec:Areas-related-to}}

Several different problem domains are either closely
related to deep transfer learning or regularly used as part of solutions. 

\paragraph{Domain adaptation\label{subsec:Domain-adaptation}}

Domain adaptation (DA) could be considered the most closely related
problem domain. In DA the task $T$ remains the same,
but the Domain $D$ differs between source and target task. According
to the definition \ref{def:transfer learning} by Pan and Yang \citep{pan2009survey}
DA falls under general transfer learning as transductive
transfer learning as described in Section \ref{sec:Overview}.
Others define it as a separate area \citep{goodfellow2016deep}. Either
way DA is not a focus of this review as it is rare
in image classification that the source and target domains have the
 same class labels. This situation is more common in natural
language processing where for example, a sentiment analysis
model could be transferred between two slightly different product review domains
\citep{wang2018deep1,zhang2017transfer}. The few
datasets that are used in DA for image classification
are carefully curated to have identical object classes in different
domains \citep{gong2012geodesic,saenko2010adapting,tommasi2014testbed}. 

Despite there being few examples of DA problems in
image classification, there are techniques that have been developed
for these tasks that could be more broadly useful to deep transfer
learning for image classification. Wang and Deng state that ``When
the labeled samples from the target domain are available in supervised
DA, soft label and metric learning are always effective'' \citep{wang2018deep1}. Using an interim fine-tuning step which incorporates metric learning on the target dataset before switching to training on the final classification task has been shown to improve performance on small fine-grained target datasets  \citep{ridnik2020tresnet}. 

\paragraph{Concept drift and multitask learning\label{subsec:Concept-drift-and multitask}}

The concept drift problem is related to DA in that
it deals with adapting models to distributions that change over time.
Multitask learning is another related problem domain where the focus
is on learning one model that can perform well at multiple tasks with different but related target datasets. While concept drift or multitask learning can be seen as types of transfer learning they have a different aim to the usual transfer learning scenario. That aim is to perform well on more than one task, rather than only the final target task.  
The major practical difference between general transfer learning and concept drift or multitask learning is that usually catastrophic forgetting is not a consideration in the former. 

\paragraph{K-shot or few shot learning\label{subsec:K-shot-or-few}}

The K-shot (also referred to as few shot) learning problem domain is specifically
focused on learning from very few or even zero labeled examples. This can be considered a specialized sub area of transfer learning and DA. 
The focus of new work in this domain is generally on improving metrics
for defining and learning the best embedding spaces and comparisons
to new examples, plus new data augmentation techniques. Transfer learning
is implicit in most of the solutions, as weights are pretrained
on a related task \citep{wang2020generalizing,kaya2019deep}.

While there is a lot of similarity, K-shot learning methods focus only on very small
target datasets. This means they do not consider how differences in the size
of the target dataset affects the optimal transfer learning methods and hyperparameters.

\paragraph{Self-supervised learning }

In machine learning, unsupervised learning refers to solutions used when there are no labeled training examples. In deep learning, we define self-supervised learning as creating target labels from a dataset without training labels. This allows the use of supervised learning methods. An aspect of the data itself is used as the training signal. The most general example is autoencoders which consist of an encoder and decoder with the target being the same as the input. The middle encoding is regularized to encourage a useful semantic encoding. Other examples
of task specific self-supervised learning include predicting the next
frame of a video \citep{srivastava2015unsupervised}, predicting the
next word in a sentence \citep{brown2020language}, image inpainting or
upsampling \citep{pathak2016context, ledig2017photo}, generative models \citep{radford2015unsupervised, goodfellow2014generative} and many more. For a detailed treatment of self-supervised methods for learning image representations see \citep{jing2020self}. 

Self-supervised learning relates closely to transfer learning as it
is often used for pretraining when there is limited, closely related labeled data. The intent is that useful relationships in the data can be learned with self-supervised pretraining before fine-tuning on the labeled data.

\subsection{Datasets commonly used in transfer learning for image classification \label{datasets}}

We anticipate that providing a comprehensive description of common source and target datasets will assist the reader in finding the best datasets for their needs. 

\paragraph{General}

General image classification datasets contain a variety of classes with a mixture of superordinate and subordinate classes from many different categories in WordNet \cite{miller1995wordnet}. ImageNet
is a canonical example of a general image classification dataset.

\paragraph{Fine-grained}
Fine-grained image classification datasets include classes with subtle differences between them for example, makes and models of cars \citep{KrauseStarkDengFei-Fei_3DRR2013} and aircraft \citep{maji13fine-grained}, and breeds of cats and dogs \citep{parkhi2012cats}.

\paragraph{Scenes}

Scene datasets contain examples of different indoor and/or outdoor scene settings.

Commonly used source datasets are described in Table \ref{source_dataset} and target datasets in Table \ref{target_dataset}.

\begin{table*}[th]

\caption{Commonly used source datasets \label{source_dataset}}
\begin{tabular}{p{5.5cm}p{2cm}p{3.8cm}p{3cm}}
\hline 
Dataset  & Size/s &  Category & Number of classes\tabularnewline
 
\hline 
ImageNet  \cite{imagenet_cvpr09} & 1.3 million+ & general & 1,000 - 21,000 \tabularnewline

JFT \cite{hinton2015distilling} & 300 million & general & 18,291 \tabularnewline

Instagram Hashtag \cite{mahajan2018exploring} & 3.5 billion & general & 1,500+ \tabularnewline

Places365 \cite{zhou2017places} & 1.8 - 8 million & scenes & 365 \tabularnewline

Inaturalist \cite{van2018inaturalist} & 859,000 & plants and animals & 5,000+ \tabularnewline

\hline 
\end{tabular}

\end{table*}

\begin{table*}[th]

\caption{Commonly used target datasets \label{target_dataset}}
\begin{tabular}{p{5.5cm}p{2cm}p{3.8cm}p{3cm}}
\hline 
Dataset  & Size/s &  Category & Number of classes\tabularnewline
 
\hline 
CIFAR \cite{krizhevsky2009learning} & 50,000 & general & 10 and 100 \tabularnewline

PASCAL VOC 2007  \cite{pascal-voc-2007} & 9,063 & general & 20 \tabularnewline

Caltech \cite{fei2004learning, griffin2007caltech} & 9,146 & general & 101 and 250 \tabularnewline

Food-101 (Food)  \cite{bossard14}& 101,000 & food objects & 500 \tabularnewline

Birdsnap (Birds) \cite{berg2014birdsnap} & 49,829 & species of birds & 500 \tabularnewline

Stanford Cars (Cars) \cite{KrauseStarkDengFei-Fei_3DRR2013} & 16,185 & models of cars & 196 \tabularnewline

FGVC Aircraft (Aircraft) \cite{maji13fine-grained} & 10,000 & models of aircraft & 100 \tabularnewline

Oxford-IIIT Pets (Pets) \cite{parkhi2012cats} & 7,049 & breeds of cats and dogs & 37 \tabularnewline

Oxford 102 Flowers (Flowers) \cite{nilsback2008automated} & 8,189 & types of flowers & 102 \tabularnewline

Caltech-uscd Birds (CUB)  \cite{wah2011caltech} & 11,788 & species of birds & 200 \tabularnewline

Stanford Dogs (Dogs) \cite{khosla2011novel} & 20,580 & breeds of dogs & 120 \tabularnewline

SUN397  SUN \cite{xiao2010sun} &  108,754 & scenes & 397 \tabularnewline

MIT 67 Indoor Scenes  \cite{quattoni2009recognizing} & 15,620 & indoor scenes & 67 \tabularnewline

Describable Textures DTD \cite{cimpoi2014describing} & 3,760 & texture adjectives & 47 \tabularnewline

Daimler Pedestrian \cite{munder2006experimental} & 23,520 & pedestrians & 2 \tabularnewline

German Road Signs \cite{stallkamp2012man} & 39,209 & road signs & 43 \tabularnewline

Omniglot \cite{lake2015human} & 1.2 million & handwritten characters & 1,623 \tabularnewline

SVHN Digits in the Wild \cite{netzer2011reading} & 73,257 & street view digits & 10 \tabularnewline

UCF101 Dynamic Images \cite{soomro2012ucf101} & 9,537 & actions & 101 \tabularnewline

\hline 
\end{tabular}

\end{table*}

 \paragraph{The Transferability Estimation benchmark} \cite{bolya2021scalable}  A benchmark for testing transferability measures. It consists of six target datasets, eight source datasets and four architectures. The target datasets include
Caltech101, CIFAR10, Birds, CUB, Oxford Pets and Stanford Dogs. The source datasets include the six target datasets plus ImageNet 1k and VOC2007. The model architectures include ResNet 50 and 18 \cite{he2016deep}, GoogLeNet \cite{szegedy2015going} and AlexNet \cite{krizhevsky2012imagenet}. 

\paragraph{Visual PETL Benchmark \cite{xin2024v}}
A benchmark for testing vision based parameter-efficient transfer learning (PETL) methods. It consists of 30 diverse, challenging, and comprehensive datasets from image recognition, video action recognition, and dense prediction tasks.

\paragraph{ Visual Decathlon Challenge (Decathlon) \cite{rebuffi2017learning}}
A challenge designed to simultaneously test performance on 10 image classification tasks being: ImageNet, CIFAR-100, Aircraft, Daimler
pedestrian, Describable textures, German traffic signs, Omniglot,
SVHN, UCF101, VGG-Flowers. All images resized to low resolution with a shorter side of 72 pixels.

\subsection{Real-world applications of deep transfer learning for image classification}

Most real world image classification data does not have the millions or even billions of independent labeled examples needed to train a complex deep learning model well from random initialization. For this reason transfer learning is an essential part of a broad range of applications. Some examples of where transfer learning is applied are highlighted below, but there are many more: 
\begin{enumerate}
    \item \textbf{Medical images.} DTLIC is often applied in the medical image domain as datasets are generally small. There are two major concerns in this domain: 
\begin{enumerate}
    \item Medical images are very different in style to image domains where data is abundant, so large, closely related source datasets are usually unavailable.
    \item Even within the same task in the medical image domain, there are differences stemming from the imaging machine and types of patient populations. These differences can mean that models do not generalize well to new data and must be fine-tuned on a limited amount of data that more closely matches the target data. 
\end{enumerate}

See \cite{kora2022transfer, kumar2024medical, juodelyte2024dataset} for a detailed overview of transfer learning and transferability in the medical image domain.

\item \textbf{Robotics.} It is more common for image classification to be part of a pipeline or backbone of a model in vision-based robotics rather than the end task. For instance, classification of objects and scene recognition can be important for navigation tasks. Action tasks can also be encoded as image classification, by encoding directions as discrete classes. Transfer learning is common in this domain, as the differences between the source and target data can come from three elements: changes to robots, tasks, or environments. If any of these items change significantly, it is likely that the model will not generalize to the new data and will need to be fine-tuned to achieve optimal performance. When the model does need fine-tuning, the aim is to generally use limited data as real world robots can be expensive and time consuming to run. While simulation environments are usually cheap and efficient for creating large amounts of data, they are generally not the final target environment, so they are far more likely to be used as the source task rather than the target task. See \cite{jaquier2025transfer} for a review of transfer learning for robotics.   

\item \textbf{Environmental modeling}.  When we consider the format of image data as input into a deep learning model we can broaden our understanding of what an image is to include most two or three dimensional spatially related data. Viewed in this way, there are many tasks in environmental modeling that can be represented well as image classification tasks. The tasks in environmental modeling are often complex. In addition, there is often a large amount of unlabeled data available, for example, satellite imagery, but an extremely limited amount of data available for the exact target prediction task. An example of this scenario is pyrocumulonimbus (PyroCb) prediction, where satellite images and other spatial domain information are used to predict whether a fire will become a PyroCb - i.e. create its own weather system, making its spread more erratic \cite{tazi2022pyrocast}. There are huge amounts of satellite images showing fires, but these specific events are very rare and complex to predict. For more details on this area  \cite{yu2025survey} discusses transfer learning techniques throughout the review.  

\item \textbf{Agriculture.} Image classification tasks are seen throughout the agriculture domain, including classifying types of crops, predicting yields, finding or classifying pests, and many more. In many of these tasks there is large amounts of unlabeled data available, but extremely limited data for the exact prediction task, for example, labeled images of a particular type of crop pest. For a more detailed discussion of deep transfer learning for image classification in this area see \cite{hossen2025transfer, olorunfemi2024advancements}.

\end{enumerate}

\section{Instance Based Approaches \label{sec:inst_appr}}
In the realm of DTLIC instance based approaches look at: 
\begin{itemize}
    \item what source data should be used for pretraining and fine-tuning
    \item whether particular source data instances or classes should be weighted more heavily during training. 
\end{itemize} 

In general, increasing the size of the source dataset increases the
performance on the target dataset. This occurs even when the target dataset is large such as ImageNet 1K (1K classes, 1.3M training images) and ImageNet 5k (5K classes, 6.6M training images) or 9k (9K classes 10.5M training images) \citep{ngiam2018domain,mahajan2018exploring,kolesnikov2020big,liu2022improved}. However, pretraining on source data that, naturally or through careful curation, matches the target data more closely often performs better than pretraining on larger, more general source datasets \citep{ngiam2018domain,mahajan2018exploring, xuhong2018explicit, sabatelli2018deep}. 

The results listed above also apply to tasks related to image classification such as object detection and semantic segmentation \cite{li2019analysis, chen2018encoder, gao2022large,mo2022review, zoph2020rethinking}

\subsection{Self-supervised pretraining techniques}

Self-supervised learning (SSL) techniques can be divided into two categories. invariant SSL (I-SSL) and equivariant SSL (E-SSL). The aim of I-SSL is to encourage the representation to be invariant to input augmentations (e.g., color jittering). Contrastive learning (CL) as the dominant invariant representation learning method, utilizes various
augmentations, including random crop, horizontal
flip, color jitter, and Gaussian blur. The objective is to embed semantically the same inputs
obtained through augmentations close together and different inputs further apart. This is based on the concept that differences due to these augmentations are inconsequential on a semantic level. E-SSL learns representations that are sensitive to the applied transformations. Many older SSL methods are equivariant, for example, RotNet \cite{komodakis2018unsupervised}, which predicts the rotation angle of an image, Jigsaw \cite{noroozi2016unsupervised} that arranges pieces of an image into the correct ordering, and Relative Patch Location \cite{doersch2015unsupervised} which predicts the relative location of two patches from the same image.  

 Recently, invariant representations have been shown to lose augmentation-related information, for example, color information, which hinders their performance on target tasks \cite{lee2021improving, gupta2023structuring}. This has resulted in more focus on E-SSL algorithms that learn features that are augmentation aware. \cite{wang2024understanding, yu2024self}. In many cases, algorithms that combine I-SSL with E-SSL techniques or I-SSL with an auxiliary loss term that predicts augmentations have achieved better performance than pure I-SSL or E-SSL techniques \cite{yu2024self, lee2021improving, wang2024understanding}.

 As an alternative E-SSL approach, recent work has focused on creating an effective tokenizer of meaningful chunks for images, equivalent to word tokenizers that have been so successful when training large language models \cite{baobeit, ramesh2021zero}. Ramesh et al. created a codebook by training a variational autoencoder (VAE) \cite{kingmaauto} on the unlabeled dataset \cite{ramesh2021zero}. This technique has been used successfully in many more recent self-supervised learning algorithms designed for image datasets.

Once the data and masked encoding task is properly encoded finer details of the task and training regime also have a significant impact on the final result  \cite{he2022masked}. Considerations include: 
\begin{itemize}
    \item \textbf{Task difficulty} significantly affects the target task performance, particularly when the pretrained weights are frozen when training on the target task. For example, when the task was masked prediction, the ideal masking ratio was shown to be higher than would be intuitive at 75\% \cite{he2022masked}. 
    \item  \textbf{Pretraining epochs}. More pretraining epochs tend to improve target task performance \cite{he2022masked}. 

\end{itemize}

Features learned from tasks incorporating tokenization tend to show more diversity in practice and benefit target datasets requiring fine-grained semantics \cite{wang2024understanding}.

There is a wide range of self-supervised learning algorithms. While algorithms that encourage some form of equivarience tend to result in better performance on the target task in many cases, there are many variations with different strengths and weaknesses. For this reason, it is often better to use a combination of different self-supervised learning strategies \cite{huang2023self}. See \cite{gui2024survey} for a detailed description of different supervised learning algorithms.

\subsubsection{Self-training}
In the absence of a much larger, closely related source dataset, self-supervised pretraining on the target dataset (self-training) before supervised fine-tuning can perform significantly better than pretraining on a less related much larger source dataset \cite{el2021large, he2022masked}. El-Nouby et al. \cite{el2021large} have shown that self-supervised learning on smaller target datasets can also be effective. Their self-training method was more effective than pretraining on ImageNet 1K for the COCO object detection target task \citep{lin2014microsoft}, and comparably effective or marginally worse for a range of small image classification target tasks very different from ImageNet 1k. Further research is required to determine the limits of the self-training regime's effectiveness, specifically in terms of the target dataset's size and the extent to which it differs from any available source dataset, to establish whether self-training on the target dataset alone is the most suitable option.

\subsection{Ranking of source datasets \label{ranking}}
All things being equal, more source data for pretraining is better than less, and a source dataset more similar to the target dataset is better. When the source dataset is similar enough labeled data is preferred to unlabeled data and uncurated or psuedo labels are generally preferred to no labels. However, when the source dataset is less similar self-supervised learning is likely to perform better than supervised learning. Recent work has attempted to bridge this gap by allowing for multimodal class distributions by enforcing similarity in the feature space between most, not all, data points with the same class label \citep{fengrethinking}. However, when the target dataset is less similar to the source dataset self-supervised pretraining is still marginally better. 

The question then becomes what should be preferred when trade-offs exist between these items. This list ranks the expected performance for types of source datasets and associated techniques: 
\begin{enumerate}
    \item Supervised training on a very similar labeled dataset that is orders of magnitude larger than the target dataset. All things being equal supervised pretraining generally performs best. 
    \item Supervised training on a closely related uncurated or psuedo labeled dataset that is orders of magnitude larger than the target dataset. If formal labels are not available, self-supervised learning is generally improved using any type of informal labels. Image tags taken from uncurated databases such as Instagram and Flickr are examples of informal labels.  Training with pseudo labels created by a collection of task-specific expert models can significantly improve performance on a target task compared to self-supervised training or training with uncurated image tags \cite{salehi2023clip}.
    \item Self-supervised training on a closely related dataset with no labels that is orders of magnitude larger than the target dataset. When there is no large labeled dataset available, self-supervised pretraining on a large unlabeled source dataset will likely perform better than using an order of magnitude smaller labeled dataset or a less related dataset \citep{zoph2020rethinking,azizi2021big}. 
    \item Self-training on the target dataset. Where there is no closely related source data, self-training (pretraining on the target dataset) is preferred to pretraining on a less closely related source dataset \cite{el2021large, he2022masked}. 
    \item Self-supervised pretraining on a less closely related source dataset. When the source dataset is less closely related self-supervised pretraining is likely to outperform supervised pretraining as it will not overfit the less related dataset.
    \item Supervised training on a less closely related target dataset. 
\end{enumerate}
The above list does not include multi-step fine-tuning processes, which, if well chosen, are likely to improve performance compared to single step processes. For example, self-supervised pretraining on a closely related source dataset that is many orders of magnitude larger than the target dataset then fine-tuning on an interim dataset that is less than an order of magnitude larger would be expected to result in better performance than simple pretraining on a closely related labeled dataset that is only one order of magnitude larger \citep{gonthier2020analysis, ng2015deep, azizi2021big, reed2022self}.

\subsection{Using dataset similarity measures in transfer learning}

Many studies have used dataset similarity or other measures to improve the results of transfer learning. Dataset similarity measures have been effectively used to increase performance on the target dataset by finding the most similar source dataset \citep{cui2018large}, source dataset classes \citep{ge2017borrowing, liu2022improved, ngiam2018domain}, or individual samples. Reinforcement learning methods have been used to learn to weight samples from the source dataset for effective pretraining \citep{zhu2020learning, yoon2020data}. Finally, dataset similarity and transferability measures have also been used successfully to pick the most effective pretrained model for a given target task \citep{you2021logme, you2022ranking, puigcerverscalable}. Transferability methods are discussed further in Section \ref{trans_meas}. The success of all these techniques for effectively using the most similar source data further showcases the importance of considering the similarity of the source and target datasets when pretraining and fine-tuning.

\subsection{Large similar target datasets}

All things being equal, pretraining on more source data is better, even when the target task is large, such as ImageNet 1K \citep{sun2017revisiting, mahajan2018exploring, dehghani2023scaling}. This is particularly so when the target task is quite different to the source task, for example transferring from an image classification task to an object detection task \cite{sun2017revisiting}.

However,  additional pretraining data is useful only if it is well correlated to the target task. \citep{huh2016makes, mahajan2018exploring}. In some cases, adding additional unrelated training data can hinder performance.  

Multi-step fine-tuning pipelines that start with pre-training on a large, less related dataset, then fine-tune on one or more smaller, more closely related datasets, before finally fine-tuning on the target task, are often more effective than single-step fine-tuning procedures \citep{yalniz2019billion, xie2020self}. Incorporating pseudo labeling into these pipelines has also been shown to be beneficial \cite{xie2020self}, particularly for target tasks specifically designed to be more difficult, such as ImageNet A, B, and P \citep{hendrycks2019benchmarking, hendrycks2021natural}.

The size of the source dataset has become even more important recently with the introduction of vision transformers with orders of magnitude more parameters than traditional CNNs \citep{dosovitskiy2020image,yu2022coca}. Large vision transformers, or a combination of transformers and CNNs, have shown superior performance to stand-alone CNNs in cases where the target dataset is large, for example, ImageNet1K, and the source dataset is orders of magnitude larger \citep{dehghani2023scaling,chen2023symbolic,yu2022coca}. 

The introduction of self-training techniques has emphasized the importance of pretraining data being closely related to the target dataset to improve performance \cite{el2021large, he2022masked}. Self-training methods can be particularly effective when applied to the largest models, such as vision transformers \cite{el2021large}.

\subsection{Large target datasets with less similar tasks \label{large_less}}

There are very few large image datasets that are not closely related
to the image classification tasks that are usually used for pretraining.
The Places365 with 1.8 million training images was used as a source
task in \citep{mahajan2018exploring}. It was shown that with less
related source and target datasets the bigger and more diverse the
source training dataset the better the results on the target dataset. It has also been shown that self-supervised pretraining can be better than supervised pretraining when the source dataset is less related \citep{azizi2021big, fengrethinking}.

Raghu et al. \citep{raghu2019transfusion} showed that when medical
image datasets are large (around two hundred thousand training examples),
pretraining on ImageNet 1K results in limited improvements over random
initialization. When the number of training examples was reduced to a small
dataset size of 5,000, pretraining with ImageNet 1K resulted in small
improvements over random initialization.

There have been mixed results shown on whether pretraining with ImageNet 1K and other common
source datasets improves object detection performance on the COCO dataset \citep{lin2014microsoft} which has 328,000 training images and 1.5 million bounding boxes across 80 classes. In recent years, several larger object detection datasets have been introduced \cite{kuznetsova2020open, shao2019objects365, cai2022bigdetection}. Shao et al. \cite{shao2019objects365} introduced the object detection dataset Objects365 which contains 365 object categories, 638 thousand images, and 10 million bounding boxes. The authors found that for the target task COCO, pretraining on Objects365 was more effective than pretraining on ImageNet 1K and resulted in a significantly shorter fine-tuning time. Additionally, the multi-step pretraining process of pretraining on ImageNet 1K, then Objects365 before transferring to the COCO target dataset slightly improved results over a single pretraining step. 

\subsection{Smaller target datasets with similar tasks}

Similarly to large target datasets, better matched pretraining data has
been shown to produce better performance on the target task than more
pretraining data \cite{ngiam2018domain, puigcerverscalable, xuhong2018explicit, cui2018large, liu2022improved, sabatelli2018deep, you2020co}. 

When pretraining deep neural networks on ImageNet 1K \citep{imagenet_cvpr09}
and transferring to significantly smaller target datasets, the improvement
of DTL over random initialization correlates positively
with the similarity of the source dataset to the target dataset.
The improvement correlates negatively with the size of the target
dataset. Both correlations are seen strongly in \citep{kornblith2019better}, with the positive correlation of performance improvement using transfer learning with the similarity between source and target datasets being most obvious. The authors state that on Stanford Cars and FGVC Aircraft, the improvement was unexpectedly small. The improvement over equivalent models trained from random initialization is marginal for these two datasets, at 0.6\% and 0.2\%. Stanford Cars
and FGVC Aircraft both contain fine-grained makes and models of cars
and aircraft respectively, whereas ImageNet 1K contains no makes and
models, just a few coarse categories for both. This means that the similarity between the source and target datasets for these two tasks is much lower than, for example, the more general CIFAR-10/100 or Oxford-IIIT Pets for which there are actually slightly more fine-grained classes in the source than the target dataset.  

It is interesting to note that the improvement in using fine-tuning over fixed pretrained features for Stanford Cars and FGVC Aircraft is significantly larger, at 25.4\% and 25.7\%, than for
any of the 10 other datasets used in the experiments. This indicates
that the pretrained features are not well suited to the task. For
comparison, at the other end of the similarity scale, Oxford-IIIT Pets shows one of the largest improvements
from 83.2\% accuracy for training from random initialization to 94.5\% accuracy for fine-tuning a pretrained model. For this dataset the increase in performance using fine-tuning compared to fixed pretrained weights is marginal at 1.1\%. This shows that the pretrained features are well suited to the task without fine-tuning.  Recent work has demonstrated several ways to effectively predict how well a pretrained model will transfer to a particular target task. These methods are covered in Section \ref{sec:best_prac}.

For smaller target datasets, large transformer models perform well where the source dataset is large and closely related to the target dataset, for example, when transferring from ImageNet 1K or other large general datasets to smaller general datasets such as CIFAR 10 and 100 \citep{dosovitskiy2020image,ridnik2023ml, zhou2021convnets}. They have not achieved performance comparable to the state of the art for tasks such as for Stanford Cars and FGVC Aircraft \citep{ridnik2023ml, liu2023learn} where no closely related large source datasets currently exist. When transformers do perform better they have been shown to have more robust features \citep{mauricio2023comparing}.

\subsection{Smaller target datasets with less similar tasks\label{subsec:Smaller-target-less-similar}}

Transfer learning has been shown to be effective in many areas where the target datasets are small and less related to ImageNet 1K and
other common source datasets. However, there have also been several
recent results that fit into this category where DTL
has shown little or no improvement over random initialization \citep{zoph2020rethinking,he2018rethinking,raghu2019transfusion}.
Transfer learning shows better performance on smaller target datasets
that are more closely related to the source dataset than larger less
related datasets in general see Section \ref{sec:best_prac} and \citep{kornblith2019better}.  Self-supervised
learning methods applied to source datasets that are more closely
related but unlabeled, or even self-training on the target task, often performs better than supervised learning
methods applied to less closely related source datasets \citep{zoph2020rethinking,azizi2021big,dai2024adaptive}.
However, even when the target dataset is very dissimilar
to the source dataset and transfer learning brings no performance
gain it can accelerate the convergence speed \citep{he2018rethinking,raghu2019transfusion}.
 
Multi-step fine-tuning pipelines can be particularly effective in the case of small target datasets with limited closely related data \citep{gonthier2020analysis, azizi2021big}, as can self-training \cite{el2021large}.

\section{Feature Based Approaches \label{sec:fea_appr}}
As discussed in Section Section \ref{sec:Overview} the deep learning interpretation of feature based approaches differs from the traditional machine learning interpretation because the feature space is learned in the former. This means that feature based approaches for DTLIC focus on creating either a shared feature space or ensuring the source and target feature spaces are as similar as possible. 

A deep learning model can be trained with a metric, that measures the distance between the source and target representations, combined with the standard cross entropy classification loss. This is to achieve the dual aims of a domain invariant model that is also a strong classifier \citep{tzeng2014deep, long2015learning}. A commonly used metric is the Maximum Mean Discrepancy (MMD)  \citep{borgwardt2006integrating} and its extensions such as conditional (MMD) \citep{long2015learning} and local (MMD) \citep{zhu2020deep}. Zhu et al. \citep{zhu2019multi} successfully extend this approach to multiple representations of the same image. 

Generative adversarial techniques are used in several feature based approaches, including: 
\begin{enumerate}
    \item To promote similarity between the source and target domain representations \citep{tzeng2017adversarial,liu2016coupled}. This approach is used most often in domain transfer to create a shared source and target domain encoding space. A deep learning model maps examples from the source and target domains to a shared feature space. An adversarial loss is used to classify whether the encoded features belong to the source or target domain. This approach performs well when no target labels are available, but it is generally less efficient than using a predefined metric or target labels when they are available. 
    \item To map target images into the source domain or vice versa \citep{taigman2017unsupervised, gamrian2019transfer}. These techniques aim to overcome differences in the input distributions $P(X)$ between the source and target datasets by learning a mapping from one to the other using an adversarial loss. They have been shown to be very effective at handling known domain shifts in images, for example, by denoising noisy images from the target dataset prior to classification \citep{gamrian2019transfer}. 
\end{enumerate}

\section{Parameter Based Approaches \label{sec:param_appr}}
\subsection{Parameter Efficient Transfer Learning}

Large pre-trained vision (VMs) and vision-language models (VLMs) have achieved impressive results in an extensive range of image classification target tasks. Full fine-tuning, though effective, still requires substantial computational and memory resources. This is particularly costly for the largest visual-language models that have hundreds of billions of parameters, but is also becoming a problem for the largest vision only models that have tens of billions of parameters. In addition, when the model is fine-tuned for a new target task, each task requires a complete set of parameters to be stored, which can quickly become costly for multiple tasks \cite{zhu2024awt, xin2024v}.

Parameter-efficient transfer learning (PETL) methods were originally designed for adapting large language models to target tasks where more limited data is available. PETL methods seek to achieve a balance between the number of parameters trained in the fine-tuning stage, and thus efficiency, and performance. Aside from the benefits in efficiency and performance PETL methods have recently been shown to be far less likely to allow memorization and subsequent leaking of fine-tuning trainng data. 

In the past two years, many PETL techniques have been developed for adapting pre-trained vision only or visual language models to downstream tasks \cite{zhong2024convolution, yu2024visual, xin2024v, shen2024expanding}. These methods can be grouped into four categories of approaches: 
\begin{enumerate}
    \item Prompt based approaches augment the input data with task-specific prompts. Visual Prompt Tuning (VPT) adapts this technique to visual transformers by prepending a set of soft prompts to the input tokens of each Transformer layer and only optimizes the prompts during fine-tuning \cite{jia2022visual}.
    \item Adapter based approaches introduce additional trainable parameters, for example,
a fully connected layer block, to the frozen pre-trained model \cite{lialin2023scaling}. 
    \item Direct selective parameter tuning (DSPT) selectively updates a subset of parameters of the pre-trained model as a trade-off between full fine-tuning and linear probing. Examples are updating just the bias terms \cite{zaken2021bitfit}, just the layer normalisation parameters, \cite{basu2024strong}, or a combination of both \cite{xie2023difffit}.
\item Efficient selective parameter tuning (ESPT) methods learn additive residuals to the original parameters, for example $\Delta W$ to update $W$. By injecting a low-rank constraint to the residuals, this category effectively reduces the learnable parameters \cite{hu2022lora}.
\end{enumerate}

For a more indepth discussion of visual PETL methods see \cite{mai2024lessons, han2024parameter}.

\subsection{Regularization Based Approaches \label{sec:reg-approaches}}

Most regularization based techniques aim to prevent fine-tuned weights from overfitting the unreliable empirical loss approximation $R(f_{n})$ \ref{subsec:Unreliable-Empirical-Risk}
by restricting the model weights or the features produced by them. They achieve this by adding a regularization term $\lambda\cdot\Omega\left(.\right)$ to the loss function: 

\begin{align}
min_{w}L\left(w\right)=\left\{ \frac{1}{n}\sum_{i=1}^{n}L\left(z\left(x_{i},w\right),y_{i}\right)+\lambda\cdot\Omega\left(.\right)\right\} 
\end{align}
\noindent
with the first term $\frac{1}{n}\sum_{i=1}^{n}L\left(z\left(x_{i},w\right),y_{i}\right)$
being the empirical loss and the second term being the regularization
term. The tuning parameter $\lambda>0$ balances the trade-off between
the two.

The success of regularization based techniques for DTL rely heavily on the assumption that the source and target datasets are closely related. This assumption implies that the optimal weights or features for the target dataset are not far from those trained on the source dataset.

There are two major categories of regularization, weight regularization and knowledge distillation.

\subsubsection{Weight regularization} 
Weight regularization directly restricts how much the model weights
can move. L2-SP regularization $\Omega(w)=\left\Vert w-w^{0}\right\Vert _{2}^{2}$
is the squared L2 norm between the source weights and the current weights \citep{xuhong2018explicit,li2020baseline}. L2SP regularization has been shown to be significantly more effective than decaying towards zero when the source dataset is very similar to the target dataset, but can harm performance when the target dataset is less similar \citep{li2020rethinking,wan2019towards,plested2021rethinking,chen2019catastrophic}. More recent work has shown that in some cases using L2-SP regularization for lower layers and L2 regularization for higher layers can improve performance compared to L2-SP or L2 regularization alone \citep{plested2021rethinking}.

\subsubsection{Knowledge distillation} 
Knowledge distillation, or feature based regularization, uses the distance
between the feature maps output from one or more layers of the source
and target networks to regularize the model:

\begin{align}
\Omega(w,w_{s})=\frac{1}{n}\sum_{j=1}^{N}\sum_{i=1}^{n}d\left(F_{j}\left(w_{t},x_{i}\right),F_{j}\left(w_{s},x_{i}\right)\right)
\end{align}
\noindent
where $F_{j}\left(w_{t},x_{i}\right)$ is the feature map output by
the jth filter in the target network defined by weights $w_{t}$ for
input value $x_{i}$, and $d\left(.\right)$ is a measure of dissimilarity
between two feature maps.

DELTA \citep{li2019delta} is an example of knowledge
distillation or feature map based regularization. It is based on the idea of retraining CNN channels that are not useful to the target task while not changing channels that are useful. Training on the target task is regularized by the attention weighted L2 loss between the final layer feature maps of the source and target models: 
\begin{align}
\Omega & (w,w^{0},x_{i},y_{i}) = \sum_{j=1}^{N}(W_{j}(w^{0},x_{i},y_{i})  \nonumber \\
& \cdot\left\Vert FM_{j}(w,x_{i})-FM_{j}(w^{0},x_{i}))\right\Vert _{2}^{2}
\end{align}
\noindent
where $FM_{j}(w,x_{i})$ is the output from the $jth$ filter applied
to the $ith$ input. The attention weights $W_{j}$ for each filter
are calculated by removing the model's filters one by one (setting
its output weights to 0), and calculating the increase in loss. Filters
resulting in a high increase in loss are then set with a higher weight
for regularization, encouraging them to stay similar to those trained
on the source task. Others that are not as useful in the target task
are less regularized and can change more. DELTA is shown to be slightly better than L2SP at improving performance on target datasets that are very similar to the source dataset, but again can decrease performance when the target dataset is less similar. \citep{kou2020stochastic}.

Attentive feature alignment (AFA) \citep{xie2021towards} improves on DELTA by calculating attention weights in the spatial domain as well as the channel level.  It adds a second term to DELTA regularization that has spatial attention weights predicted by a two layer neural network: 

\begin{align}
\Omega_C & (w,w^{0},W^s_a, x_{i}) = \sum_{j=1}^{Height}\sum_{k=1}^{Width}  \nonumber \\
& \cdot\left\Vert A^i_{j,k} FM_{j}(w,x_{i})-FM_{j}(w^{0},x_{i}))\right\Vert _{2}^{2}
\end{align}
\noindent

Where height and width are the height and width of each feature map and $A^i_{j,k}$ is the attention weight at that position calculated by the neural network. 
AFA is shown to improve on L2-SP and DELTA when the source and target datasets are very similar. 

Wan et al. \citep{wan2019towards} propose a method that prevents the regularization term from moving the weights in the opposite direction of the empirical loss term in the hope that this will stop negative transfer.
They show that their proposal slightly improves performance with a
ResNet-18 model on four different target datasets. 

Batch spectral shrinkage (BSS) \citep{chen2019catastrophic} introduces
a loss penalty applied to smaller singular values of channelwise features
in each batch update during fine-tuning so that untransferable spectral
components are suppressed. The results show that their method never hurts
performance on the given datasets and often produces significant performance
gains over L2, L2-SP and DELTA regularization for smaller target datasets. They also show that BSS can improve performance for less similar target datasets where L2-SP hinders performance. 
Sample-based regularization \citep{jeon2020sample} proposes regularization
using the distance between feature maps of pairs of inputs in the
same class, as well as weight regularization.  The authors
report an improvement over L2-SP, DELTA and BSS in all tests.  Their results reconfirm that BSS performs better than DELTA and
L2SP in most cases and in some cases DELTA and L2SP decrease performance
compared to the standard L2 regularization baseline.

Given the number of methods showing only small improvements compared to previous methods, more work is needed to show comparisons across a broad range of tasks and models. In particular, most work so far has focused on how DTL regularization techniques compare when the source and target dataset are very similar. More investigation is needed to show how each of these methods perform when applied to less similar datasets. 

In some cases using L2-SP regularization
for lower layers and traditional L2 regularization decaying the weights towards zero for higher layers can improve performance over using either one for all layers \citep{plested2021rethinking}. While the evidence for this is
so far limited it does align with observations from \citep{yosinski2014transferable}
that lower layers are more transferable than higher layers.  Recent work shows a higher learning rate for lower layers is needed when the distribution shift between the source and target dataset is in the form of low level image corruptions \cite{leesurgical}. We recommend further investigation into whether more regularization should be applied to lower layers compared to higher layers in this case, too. 

\subsection{Normalization Based Approaches}

Several techniques attempt to better align fine-tuning in the target domain with the source domain by making adjustments to the standard batch normalization or other forms of normalization that are used between layers in modern deep learning models.
\begin{enumerate}
\item Sharing batch normalization hyperparameters across source and target domains has been shown to be more effective than having separate ones across many domain adaptation tasks \citep{wang2019transferable,maria2017autodial}. Wang et al. \citep{wang2019transferable} introduce an additional batch
normalization hyperparameter called domain adaptive $\alpha$.  This
takes standard batch normalization with $\gamma$ and $\beta$ shared
across source and target domain and scales them based on the transferability value of each channel calculated using the mean and variance statistics prior to normalization. These techniques have been shown to improve performance in unsupervised domain adaptation, however as far as we are aware they have
not been applied to the general supervised transfer learning case.
\item Stochastic normalization \citep{kou2020stochastic} samples batch normalization
based on mini-batch statistics or based on moving statistics for each
filter with probability hyperparameter p. At the start of fine-tuning
on the target dataset the moving statistics are initialized with those
calculated during pretraining in order to act as a regularizer. This
is designed to overcome problems with small batch sizes resulting
in noisy batch-statistics or the collapse in training associated with
using moving statistics to normalize all feature maps \citep{ioffe2017batch,ioffe2015batch}.
The results show that their methods improve over BSS, DELTA
and L2-SP for low sampling versions of three standard target datasets and improve over all but BSS for larger versions of the same datasets.
\item A more efficient training mode designed specifically for fine-tuning is proposed in \citep{you2024efficient}. The authors designed a training mode for batch normalization that they prove is equivalent to standard eval mode, but more memory and computation efficient than the standard eval mode that is usually used. Their results are also shown empirically.  

\end{enumerate}

In addition to the above, there are also techniques for generalising a model trained in one domain to either a related domain or out of distribution data. These techniques apply normalization approaches to align test data statistics with the training domain \citep{fan2021adversarially, liuncertainty, nado2020evaluating}. A full discussion of these techniques is beyond the scope of this work but can be found in \cite{huang2023normalization}.

\subsection{Ensemble Strategies}

Ensembling of different deep learning models is a well known technique to improve inference performance. When applied to transfer learning ensembling strategies can improve fine-tuning by combining more than one strategy, for example fine-tuned and frozen layers, or different fine-tuning hyperparameters. 

Guo et al. \citep{guo2019spottune} make two copies of their ResNet
models pretrained on ImageNet 1K. One model is used as a fixed feature selector with the pretrained layers frozen and the other model is fine-tuned. They reinitialize the final classification layer in both. A policy net trained with reinforcement learning is then used to create a mask to combine layers from each model together in a unique way for each target example. They report state of the art results on the decathlon 10 task dataset. Their results are less competitive on other more standard datasets like Stanford Cars and Flowers, however, they do show improvements over simple fine-tuning. A similar technique with a genetic algorithm used to select which layers to freeze or fine-tune is shown to be more effective than a simple fine-tuning strategy in \cite{nagae2022automatic}. Nagae et al. \cite{nagae2022automatic} used normalized optimal transport distance between the features of the source and target dataset for each layer to decide which layers to freeze or fine-tune. This technique showed improved accuracy on the target dataset when tested on small greyscale datasets. More testing is needed to confirm this result for a broader range of datasets \cite{nagae2022automatic}. 

MultiTune \citep{wang2020multitune} simplifies SpotTune by removing the policy network and concatenating the features from each model prior to the final classification layer. It also improves on SpotTune and \cite{nagae2022automatic} by using two different non-binary fine-tuning \citep{plested2021rethinking} hyperparameter settings rather than one fine-tuned and one frozen model. The results show that MultiTune improves or equals accuracy compared to SpotTune in most cases tested with significantly less training time. AMF: Adaptable Weight Fusion \citep{shen2022amf} further improves on MultiTune by adding a policy network that learns the optimal combination of the final features of each model, rather than simply concatenating them and results in further improved performance.  Antonio et al. achieve similar results using a variation of adaptable weight fusion that learns a weighting of two models trained on different subsets of the source data (bagging) \cite{antonio2025efficient}.

Ensembling can also be effective in pretraining, particularly when many similar but small source datasets are available. In \cite{du2024datamap}, similarity between the target and potential source datasets was calculated using cosine similarity between convolutional kernels pretrained on a large source dataset. Higher accuracy was achieved when doing ensemble pretraining using all potential source datasets over a preset threshold compared to using only one source dataset. 

\subsubsection{Ensemble models}
 In transfer learning it is usual that different models initialized with the same pretrained weights will end up in the same weight basin \cite{neyshabur2020being}. Ensembles of models with different pretrained weights (global ensemble) significantly outperform those with the same weight initialization (local ensemble), particularly when the target task is less similar to the source task \cite{sadrtdinov2023stay}. The weights from local ensemble models can be averaged to create a model soup \cite{rame2022diverse, wortsman2022model}. While model soups have been shown to have slightly inferior performance to local ensemble models on in-distribution data, they are more robust to out-of-distribution data and have significantly lower inference compute costs \cite{rame2022diverse, wortsman2022model}. Recent work has created semi-local model soups by averaging weights created from the same pretrained models with diverse fine-tuning hyperparameters. With a good combination of hyperparameters these semi-local model soups can partially bridge the gap between local ensembles/model soups and global ensembles, balancing efficiency and prediction performance \cite{sadrtdinov2023stay}.

Supervised pretraining tends to result in better performance on downstream target tasks than self-supervised learning, all things (size and similarity of the source and target datasets) being equal. However, there is some evidence that this may not hold with ensemble models \cite{sadrtdinov2023stay}. More work is needed to explore this finding. Future work should also consider the similarity between the source and target datasets and the size of the target dataset when deciding on a model ensemble or model soup scheme.

\section{Best Practice Categorization \label{sec:best_prac}}

There are many different questions that need to be examined when deciding on how to perform transfer learning for a particular source and target task. They can be broadly categorized by the following questions: 

\begin{enumerate}
    \item \textbf{What} type of solutions should be considered?
    \item \textbf{How} should the potential solutions and training data be used?
    \item \textbf{Which} hyperparameters should be used?
    \item \textbf{Are} there ways to decide whether transfer learning will be successful and how to do it?
    
\end{enumerate}

The distinction between target dataset sizes discussed in \ref{sec:domain_cat} is useful here as it has been
shown that small target datasets are much more sensitive to changes
in transfer learning settings \citep{plested2019analysis, liu2022improved}. It has
also been shown that standard transfer learning protocols do not perform as well when transferring to a less related target task \citep{he2018rethinking,kornblith2019better, plested2021rethinking},
with negative transfer being an extreme example of this \citep{wang2019characterizing,wang2019transferable},
and that the similarity between datasets should be considered when
deciding on hyperparameters \citep{li2020rethinking,plested2021rethinking}.
These distinctions go some way toward explaining conflicting performance
of DTL methods in recent years \citep{he2018rethinking,li2020rethinking,wan2019towards,zoph2020rethinking}.

\subsection{Measures of transferability \label{trans_meas}}

We start by answering the final question above - are there ways to decide whether transfer learning will be successful and how to do it? 

Although empirical transferability can be the gold standard of describing how easy it is to transfer the knowledge learned from a source task to a target task, it is computationally
expensive to obtain. A transferability metric is a
function of the source and target data that approximates the
empirical transferability. It is, therefore, imperative to find
efficient transferability metrics that can accurately estimate
empirical transferability \cite{tan2024transferability}.

Measures of transferability can show us how well the current model is fitting the new dataset and how far the weights need to be moved to find an optimal solution. This can help us to:
\begin{itemize}
    \item determine how well transferring a specific pretrained model to a new target dataset will perform
    \item determine whether negative transfer is likely
    \item potentially identify the best transfer learning protocols to use.
\end{itemize}   

Measures of transferability can be divided into Source-Free Model Transferability
Estimation (SF-MTE), which assumes no access to the source dataset, and
Source-Dependent Model Transferability
Estimation (SD-MTE), which assumes access to the source dataset \cite{ding2024model}. 

SF-MTE methods include static methods and dynamic methods. Static methods calculate scores directly based on statistical information, such as features and logits computed on the target dataset using candidate source
models. Dynamic methods compute scores using mapping functions
or learning frameworks to map the original static information into a different space. Some well known static SF-MTE methods include H-Score \citep{bao2019information}, LEEP \citep{nguyen2020leep}, negative conditional entropy (NCE) \cite{tran2019transferability} and FERM \cite{fu2022ferm}. 
\begin{enumerate}
    \item Feature structure-based methods primarily focus on the intra-class and inter-class separation of features. Commonly used methods include H-Score \citep{bao2019information}.
    \item Bayesian Statistic methods are based on estimating the joint distribution of source and target datasets. The most well known model family under these methods is LEEP \citep{nguyen2020leep}.
    \item Information theory based methods measure incorporate entropy and mutual information \cite{tran2019transferability}.
\end{enumerate}

Dynamic SF-MTE methods usually aim to approximate changes in the model during fine-tuning using an efficient transformation. Methods in this area include LogME \cite{you2021logme, you2022ranking} and KITE \cite{guo2024kite} .

SD-MTE can also be further divided into static and dynamic methods. Since the source data is available, dataset similarity methods can be incorporated. Static SD-MTE methods include many variations of the application of optimal transport distance (OT), also known as Wasserstein distance (W-distance), including OTCE, JC-NCE, F-OTCE, and JC-OTCE \cite{tan2021otce, tan2024transferability}.

Dynamic SD-MTE also involves a transformation that maps the original information into a different space. In this case using dynamic features such as gradients or learning frameworks.

The principle gradient expectation (PGE) of the source and target dataset is calculated in \cite{qi2022transferability} and their
similarity is compared to approximate the transferability.

There have been limited studies on the robustness of transferability estimators. An initial study found that minor variations in experimental setups can result in different conclusions regarding the performance of one transferability metric over another \citep{agostinelli2022stable}. Another limited study found that there are no transferability metrics capable of consistently estimating the performance on the target datasets in the medical domain \citep{chaves2023performance}.

It should be noted that measures of similarity introduced in Section \ref{sec:sim}, can also be used as a proxy for transferability, as the more similar a source and target dataset are, the more effective transferring a model pretrained on the source dataset to the target dataset will be.

\subsection{General best practice}

Transfer learning protocols have been heavily influenced by early work performed by Yosinski et al. \citep{yosinski2014transferable}. They found that when the source and target datasets are extremely similar and the target dataset is large (over 500,000 training examples), transferring more layers is better, and fine-tuning results in better performance than freezing layers. It is important to note that their results were only shown to hold for large and closely related target datasets and their
specific fine-tuning hyperparameters. In fact, they demonstrated that higher layers are less transferable when the source and target datasets are less related. More recent work has found that transferring less layers may result in better performance when the source and target datasets are less similar \cite{plested2019analysis, plested2021rethinking}, and the optimal learning rate depends on the size and similarity of the target dataset compared to the source dataset \cite{plested2021rethinking, leesurgical}.

There are some general guidelines for optimal transfer learning that do not depend on the size or similarity of the source and target datasets. These include: 
\begin{enumerate}
    \item \textbf{Fine-tuning}:  tends to work better than freezing weights in almost all cases
\citep{yosinski2014transferable, plested2019analysis, huang2023self}. To find the optimal learning rate for fine-tuning the similarity between the source and target datasets and tasks should be considered. The layers that are likely to be most useful to the target task should also be considered \cite{plested2021rethinking}. For example, Lee at el. found that a higher learning for lower layers performs better when the distribution shift is in the form of low level image corruptions \cite{leesurgical}. Recent work has shown that gradually freezing layers with low gradient norms throughout training can result in better performance than standard fine-tuning \cite{davila2023gradient}. More work should be done to confirm this result on a wider range of datasets.  
\item \textbf{More versus closely related source data}. More source and target training data is better in general.  \citep{ngiam2018domain,mahajan2018exploring,kolesnikov2020big}.
However, a smaller, closely related dataset is better than a larger, less similar dataset\citep{ngiam2018domain,mahajan2018exploring}. Pretraining with noisy, uncurated labels or psuedo labels can improve final performance over purely self-supervised training. When a large, closely related source dataset is not available, self-training can result in significantly better performance on the target task compared to pretraining on a less related source dataset \cite{el2021large, he2022masked}. See Section \ref{ranking} for a full ranking of source datasets and techniques. 

\item  \textbf{Model selection}. Model architectures that perform better on ImageNet were found to perform better on a range of target datasets in \citep{kornblith2019better}. For models trained with a wide range of self-supervised pretraining methods on a large source dataset, few shot fine-tuning performance on the source dataset has been shown to correlate closely with performance on a range of diverse downstream target tasks.  
\item \textbf{Number of layers pretrained}. The number of layers to reinitialize from random weights is strongly related to the optimal
fine-tuning learning rate.  It is likely that a larger learning rate is optimal when
more layers are pretrained and a lower learning rate when less layers
are pretrained \citep{plested2019analysis,plested2021rethinking}. For this
reason it is important to tune both the number of layers pretrained and learning rate together. Given that a higher fine-tuning learning rate on lower layers than higher layers has been shown to be optimal when the difference between the source and target dataset are low level image distortions more work should be done to see whether reinitializing lower layers instead of upper layers could be optimal for these types of distribution shifts \cite{leesurgical}.
\item  \textbf{Regularization}. L2-SP  or other more recent transfer learning specific regularization techniques like DELTA, BSS, stochastic normalization, etc. improve performance when the source and target dataset are closely related, but often hinder it when they are less
related \citep{xuhong2018explicit,li2019delta,wan2019towards,plested2021rethinking}.
These regularization techniques are discussed in more detail in Section \ref{sec:reg-approaches}.

\item \textbf{Measures of transferability}: can predict the performance of a pretrained model on a particular target task \citep{bao2019information,tran2019transferability,nguyen2020leep,you2022ranking, fu2022ferm}. The performance of a simple classification model with frozen weights
can give insight into transfer learning best practice for the target task \citep{plested2021rethinking}.   More work is needed to determine if transferability measures in general can help determine optimal transfer learning hyperparameters across a wide range of source and target datasets.

\end{enumerate}

\subsection{Best Practice for Large Similar Target Datasets}

Most techniques will work well in this case, so less time needs to be spent finding the best technique for the given task \citep{yosinski2014transferable, kolesnikov2020big}. Transferring all layers is likely to be optimal \citep{plested2019analysis, plested2021rethinking}. A lower learning rate and momentum are better for fine-tuning when the source and target datasets are similar \citep{li2020rethinking,plested2021rethinking, dai2024adaptive}. However, this may not be necessary when the target
dataset is very large and overfitting a poor empirical loss minimizer is less of a concern. If a higher learning rate is used it should be decayed quickly so as not to change the pretrained weights too much. All weights should be fine-tuned not frozen \citep{yosinski2014transferable, plested2019analysis, huang2023self}. L2-SP regularization or other more recent techniques like DELTA, BSS, stochastic normalization etc. are likely to work reasonably well \citep{li2020rethinking,wan2019towards,plested2021rethinking,chen2019catastrophic}. However, a very large target dataset may need less regularization in general.

\subsection{Best Practice for Large, Less Similar Target Datasets}

In this scenario, negative transfer is a concern as it is much more difficult to improve performance using transfer learning. Care must be taken when selecting transfer learning protocols, and self-training on only the target dataset should be considered. 

If transferring straight from a less related source dataset a higher learning rate, momentum and training for longer before decaying the learning rate should be used when fine-tuning \citep{li2020rethinking,plested2021rethinking}. This is to attempt to move weights out of the flat basin of the loss landscape that is created when using pretrained weights, and further away from their sub-optimal pretrained values \citep{neyshabur2020being, liu2019towards}. 

Recent regularization techniques like L2-SP, DELTA, BSS, stochastic normalization, etc. should not be used in this case as weights should be allowed to move freely away from their pretrained values based on the larger target dataset. Pretraining fewer layers of weights could also be attempted in order to take advantage of the more general lower layers, while allowing the more task specific higher layers to train from random initialization \citep{yosinski2014transferable, plested2021rethinking}. Alternatively fine-tuning higher layers at a higher learning rate than lower layers could take advantage of the fact that lower layers are likely to be more general and need changing less than the task specific higher layers. 

Finally, a global ensemble created from models fine-tuned from different pretrained starting points should be considered \cite{sadrtdinov2023stay}. 

\subsection{Smaller Target Dataset Considerations Overview}

It is often not possible to train deep neural networks on small datasets
from random initialization with supervised learning \citep{kora2022transfer, mormont2018comparison, kraus2017automated,heker2020joint, el2021large, he2022masked}. This means that
transfer learning becomes more important as the size of the
target dataset is reduced. It has also been shown that the transfer learning
protocols, including the number of layers to transfer and fine-tuning
hyperparameters have a much greater impact on performance as the size
of the target dataset is reduced \citep{plested2019analysis, plested2021rethinking}. As the    
size of the target dataset decreases, two competing factors affect
transfer learning performance:
\begin{enumerate}
\item The empirical loss estimate becomes less reliable, as described in Section  \ref{subsec:Unreliable-Empirical-Risk}
making overfitting idiosyncrasies in the target dataset more likely.
\item The pretrained weights implicitly regularize the fine-tuned model more strongly and the final weights generally do not move as far from their pretrained values
\citep{neyshabur2020being,liu2019towards,raghu2019transfusion}.
\end{enumerate}
The outcome of the first point is an increasing need to use transfer learning
and other methods to reduce overfitting. The implicit regularization
noted in the second point can have a positive impact in reducing
overfitting the empirical loss as per the first point. It can also
have a negative impact (negative transfer) if the weights transferred
from the source dataset are ineffective for the target dataset.
When the weights, and thus the features produced, are restricted to
being far from optimal the negative effect on performance can be
compounded by the first point as inappropriate features can be used to overfit the new task \citep{plested2019analysis}.

Perhaps counterintuitively, transferring less than all available layers is often optimal when the target dataset is small, even when the source and target datasets are extremely similar. This is because it is more difficult to change large pretrained weights based on a small target dataset and so overfitting inappropriate features becomes more likely \citep{plested2019analysis, plested2021rethinking}.

While generally accepted practice is to train higher layers at higher learning rates a higher fine-tuning learning rate on lower layers than higher layers has been shown to be optimal when the difference between the source and target dataset are low level image distortions \cite{leesurgical}.

In general,  the learning rate should be decayed quickly with smaller target datasets  \citep{plested2019analysis, kolesnikov2020big}. However, when the target data set is small it must be taken into account that the number of weight updates per epoch will be low and the number of updates should be reduced, not necessarily the number of epochs.

\subsection{Best Practice for Small Similar Target Datasets}
This scenario is where transfer learning really excels, although more care needs to be taken to use optimal hyperparameters when the dataset is small. The experiments from \citep{yosinski2014transferable} using large similar target datasets were repeated in \citep{plested2019analysis}, but included a range of smaller target dataset sizes. These experiments demonstrated that findings for large similar target datasets do not carry over to smaller target datasets. A significant improvement in performance was shown using optimal transfer learning protocols for small target datasets compared to more commonly used protocols from \citep{yosinski2014transferable}. This improvement increased considerably as the size of the target dataset
decreased. 

For small similar target datasets the optimal learning rate and momentum will likely be low as the weights will not need to be moved far from their pretrained values and if they are, overfitting is more likely with a small dataset \citep{plested2019analysis, kolesnikov2020big}. The learning rate should be decayed quickly for the same reason. 

Recent regularization techniques like L2-SP, DELTA, BSS, stochastic normalization, etc. are likely to improve performance significantly in this case \citep{li2020rethinking,wan2019towards,plested2021rethinking,chen2019catastrophic}.

\subsection{Best Practice for Small Less Similar Target Datasets}

In this case transfer learning can be very useful if done well, but can also lead to poor results. As highlighted in Section \ref{sec:reg-approaches}  there are many methods designed to improve performance in transfer learning that perform well when the target dataset is small and similar, but poorly when the target dataset is less similar \cite{nguyen2020leep, xuhong2018explicit, li2020baseline, kou2020stochastic, wan2019towards}. It is difficult to strike an optimal balance between: 
\begin{enumerate}
\item{} allowing the weights to move far enough from their pretrained values that the model is not using inappropriate features for the classification task
\item{} not allowing the weights to overfit the unreliable empirical loss.
\end{enumerate}

Given this, if there is any way to use a more closely related dataset it is likely to improve results. This could be:
\begin{enumerate}
\item{} Doing self-supervised pretraining on a large unlabeled more closely related dataset instead of supervised pretraining on a large less similar dataset. 
\item{} Using a moderately sized more closely related dataset either instead of a large source dataset for pretraining or as an interim fine-tuning step when transferring weights from a less related source task to the final target task.
\item{} Using only more related classes from the source dataset during pretraining or weighting these classes more heavily.
\end{enumerate}

If it is not possible to use a more related dataset, then it is likely that doing self-training on the target dataset before supervised learning on the final target task will perform better than pretraining on a less related source dataset.

If  a less similar source dataset is used a high initial learning rate and momentum is likely to be optimal \cite{dai2024adaptive}, but the learning rate should be decayed quickly. It may be better to fine-tune higher layers at a higher learning rate than lower layers \citep{plested2021rethinking} as the weights from lower layers tend to be more generally applicable \cite{yosinski2014transferable}. Recent regularization techniques like L2-SP, DELTA, BSS, stochastic normalization, etc. should not be used on all layers. It has been shown that L2-SP and DELTA particularly can result in minimal improvement or even negative transfer when the source and target datasets are less related \citep{li2020rethinking,wan2019towards,plested2021rethinking,chen2019catastrophic}. However, if $P(X_S)$ and  $P(X_T)$ are similar it may be beneficial to use these techniques on lower layers even if $P(Y_T\vert X_S)$ and $P(Y_T\vert X_T)$ are dissimilar.

\section{Discussion \label{sec:disc}}

There are two overarching themes throughout this review of transfer
learning techniques and applications:
\begin{enumerate}
\item More source data is better in general, but more closely related
source data for pretraining will often produce better performance
on the target task than a larger source dataset. 
\item The size of the target dataset and how closely related it is to the
source dataset strongly impacts the performance of transfer learning.
Using sub-optimal transfer learning hyperparameters can result in
negative transfer when the target dataset is less related and large
enough to be trained from random initialization. Even when the target dataset is small, self-training on the target dataset is likely to perform better than pretraining on a less related source dataset.
\end{enumerate}
Currently, there tends to be an all or nothing approach to transfer
learning. Either transferring all layers improves performance or it
does not and possibly decreases performance (negative transfer). The
same approach is often taken to freezing layers, either layers are
frozen or they are fine-tuned at the same learning rate as the rest
of the model, and again with weight regularization, either all transferred weights
are decayed towards their pretrained values (L2SP) or towards zero
(or sometimes not at all). For example, the binary way to consider whether to pretrain on ImageNet 1K when the target task is object detection is to decide whether to transfer all but the final layer of weights and minimally fine-tune them so they do not move far from their pretrained values. This has been shown to result in minimal improvement or negative transfer \citep{he2018rethinking}.

We advocate that this all or nothing binary approach
should be discarded, and instead, all decisions made about how to perform
transfer learning should be thought of as sliding scales. Transferring
all layers, freezing layers and decaying all weights towards pretrained
values are at one extreme of the scale and likely to be reasonable
only for source and target datasets that are extremely similar. Training
from random initialization is at the other end of the scale and is likely to only
be optimal if the source and target domain have no similarities. This
aligns with the observations in \citep{yosinski2014transferable}
that \textquotedblleft first-layer features appear not to be specific
to a particular dataset or task, but general in that they are applicable
to many datasets and tasks\textquotedblright . More work is needed
to show how much transfer learning is optimal for a given source and
target dataset relationship and target dataset size, rather than focusing on
whether transfer learning is effective.

\section{Conclusion \label{sec:conc}}
We presented a new deep transfer learning taxonomy for image classification. Previous transfer learning taxonomies have not considered the size of the target dataset and its similarity to the source dataset. We have shown that these two characteristics are particularly important in determining the optimal transfer learning solution for a given target task. 

As discussed in Section \ref{sec:neg_tran} negative transfer can only occur when too many weights are pretrained or transferred weights are prevented from moving far enough from their pretrained values due to inappropriate fine-tuning. If we think of transfer learning solutions in a more continuous rather than binary space we can eliminate negative transfer.  Instead of deciding whether or not to transfer at all we can consider the question of how much to transfer.  The similarity between the source and target datasets and the size of the target dataset determines how many layers should be transferred from large pretrained values that are unlikely to move much, and how many should be reinitialized to small values that are better able to change to fit the task. This similarity also determines how fine-tuning hyperparameters should be selected to ensure the pretrained weights do not change too little or too much based on the target dataset. 

In the image classification domain the similarity between source and target dataset can often be estimated using a visual assessment combined with existing knowledge. For example, the following are likely to be dissimilar:
\begin{itemize}
    \item RGB image datasets compared to monochrome
    \item fine-grained classification tasks compared to general classification tasks
    \item tasks from very different domains, such as general object classification and medical imaging tasks.
\end{itemize}
If the similarity is not obvious, measures of transferability can be used as a proxy. 

Future work should focus on solutions that take into account our new taxonomy and in particular the size of the target dataset and its similarity to the source dataset. There is a large body of work featuring solutions that are mostly effective when target datasets are small and closely related to the source dataset. We know that transfer learning is likely to perform better in this case. We recommend more focus on target tasks that have moderate to limited closely related data utilizing our guidance on eliminating negative transfer in these challenging cases.

\noindent

\bibliographystyle{plain}
\bibliography{thesis}

@inproceedings{brown2020language,
  title = {Language Models are Few-Shot Learners},
   author = {Brown, Tom and Mann, Benjamin and Ryder, Nick and Subbiah, Melanie and Kaplan, Jared D and Dhariwal, Prafulla and Neelakantan, Arvind and Shyam, Pranav and Sastry, Girish and Askell, Amanda and Agarwal, Sandhini and Herbert-Voss, Ariel and Krueger, Gretchen and Henighan, Tom and Child, Rewon and Ramesh, Aditya and Ziegler, Daniel and Wu, Jeffrey and Winter, Clemens and Hesse, Chris and Chen, Mark and Sigler, Eric and Litwin, Mateusz and Gray, Scott and Chess, Benjamin and Clark, Jack and Berner, Christopher and McCandlish, Sam and Radford, Alec and Sutskever, Ilya and Amodei, Dario},
   booktitle = {Advances in Neural Information Processing Systems},
   publisher = {Curran Associates, Inc.},
   volume = {33},
   pages = {1877–1901},
   year = {2020},
   type = {Conference Proceedings}
}

@inproceedings{srivastava2015unsupervised,
  title={Unsupervised learning of video representations using lstms},
  author={Srivastava, Nitish and Mansimov, Elman and Salakhudinov, Ruslan},
  booktitle={International conference on machine learning},
  pages={843--852},
  year={2015}
}

@inproceedings{ridnik2020tresnet,
  title = {TResNet: High Performance GPU-Dedicated Architecture},
   author = {Ridnik, Tal and Lawen, Hussam and Noy, Asaf and Ben, Emanuel and Sharir, Baruch Gilad and Friedman, Itamar},
   booktitle = {2021 IEEE Winter Conference on Applications of Computer Vision (WACV)},
   pages = {1399-1408},
   DOI = {10.1109/WACV48630.2021.00144},
   year = {2021},
   type = {Conference Proceedings}
}

@inproceedings{guo2019spottune,
  title={Spottune: transfer learning through adaptive fine-tuning},
  author={Guo, Yunhui and Shi, Honghui and Kumar, Abhishek and Grauman, Kristen and Rosing, Tajana and Feris, Rogerio},
  booktitle={Proceedings of the IEEE Conference on Computer Vision and Pattern Recognition},
  pages={4805--4814},
  year={2019}
}

@inproceedings{ge2017borrowing,
  title={Borrowing treasures from the wealthy: Deep transfer learning through selective joint fine-tuning},
  author={Ge, Weifeng and Yu, Yizhou},
  booktitle={Proceedings of the IEEE conference on computer vision and pattern recognition},
  pages={1086--1095},
  year={2017}
}

@inproceedings{plested2019analysis,
  title={An Analysis of the Interaction Between Transfer Learning Protocols in Deep Neural Networks},
  author={Plested, Jo and Gedeon, Tom},
  booktitle={International Conference on Neural Information Processing},
  pages={312--323},
  year={2019},
  organization={Springer}
}

@article{ngiam2018domain,
  title={Domain adaptive transfer learning with specialist models},
  author={Ngiam, Jiquan and Peng, Daiyi and Vasudevan, Vijay and Kornblith, Simon and Le, Quoc V and Pang, Ruoming},
  journal={arXiv preprint arXiv:1811.07056},
  year={2018}
}

@article{huh2016makes,
  title={What makes ImageNet good for transfer learning?},
  author={Huh, Minyoung and Agrawal, Pulkit and Efros, Alexei A},
  journal={arXiv preprint arXiv:1608.08614},
  year={2016}
}

@inproceedings{he2018rethinking,
  title = {Rethinking ImageNet Pre-Training},
   author = {He, Kaiming and Girshick, Ross and Dollar, Piotr},
   booktitle = {2019 IEEE/CVF International Conference on Computer Vision (ICCV)},
   pages = {4917-4926},
   DOI = {10.1109/ICCV.2019.00502},
   year = {2019},
   type = {Conference Proceedings}
}

@inproceedings{yosinski2014transferable,
  title={How transferable are features in deep neural networks?},
  author={Yosinski, Jason and Clune, Jeff and Bengio, Yoshua and Lipson, Hod},
  booktitle={Advances in neural information processing systems},
  pages={3320--3328},
  year={2014}
}

@inproceedings{sun2017revisiting,
  title={Revisiting unreasonable effectiveness of data in deep learning era},
  author={Sun, Chen and Shrivastava, Abhinav and Singh, Saurabh and Gupta, Abhinav},
  booktitle={Proceedings of the IEEE international conference on computer vision},
  pages={843--852},
  year={2017}
}

@inproceedings{mahajan2018exploring,
  title={Exploring the limits of weakly supervised pretraining},
  author={Mahajan, Dhruv and Girshick, Ross and Ramanathan, Vignesh and He, Kaiming and Paluri, Manohar and Li, Yixuan and Bharambe, Ashwin and van der Maaten, Laurens},
  booktitle={Proceedings of the European Conference on Computer Vision (ECCV)},
  pages={181--196},
  year={2018}
}

@inproceedings{cui2018large,
  title={Large scale fine-grained categorization and domain-specific transfer learning},
  author={Cui, Yin and Song, Yang and Sun, Chen and Howard, Andrew and Belongie, Serge},
  booktitle={Proceedings of the IEEE conference on computer vision and pattern recognition},
  pages={4109--4118},
  year={2018}
}

@inproceedings{krizhevsky2012imagenet,
  title={Imagenet classification with deep convolutional neural networks},
  author={Krizhevsky, Alex and Sutskever, Ilya and Hinton, Geoffrey E},
  booktitle={Advances in neural information processing systems},
  pages={1097--1105},
  year={2012}
}

@inproceedings{imagenet_cvpr09,
        AUTHOR = {Deng, J. and Dong, W. and Socher, R. and Li, L.-J. and Li, K. and Fei-Fei, L.},
        TITLE = {{ImageNet: A Large-Scale Hierarchical Image Database}},
        BOOKTITLE = {CVPR09},
        YEAR = {2009},
        BIBSOURCE = "http://www.image-net.org/papers/imagenet_cvpr09.bib"}

@inproceedings{kornblith2019better,
  title={Do better imagenet models transfer better?},
  author={Kornblith, Simon and Shlens, Jonathon and Le, Quoc V},
  booktitle={Proceedings of the IEEE Conference on Computer Vision and Pattern Recognition},
  pages={2661--2671},
  year={2019}
}

@inproceedings{bengio2015sharing,
  title={Sharing representations for long tail computer vision problems},
  author={Bengio, Samy},
  booktitle={Proceedings of the 2015 ACM on International Conference on Multimodal Interaction},
  pages={1--1},
  year={2015}
}

@article{wang2020generalizing,
  title={Generalizing from a few examples: A survey on few-shot learning},
  author={Wang, Yaqing and Yao, Quanming and Kwok, James T and Ni, Lionel M},
  journal={ACM Computing Surveys (CSUR)},
  volume={53},
  number={3},
  pages={1--34},
  year={2020},
  publisher={ACM New York, NY, USA}
}

@article{zhuang2020comprehensive,
  title={A comprehensive survey on transfer learning},
  author={Zhuang, Fuzhen and Qi, Zhiyuan and Duan, Keyu and Xi, Dongbo and Zhu, Yongchun and Zhu, Hengshu and Xiong, Hui and He, Qing},
  journal={Proceedings of the IEEE},
  year={2020},
  publisher={IEEE}
}

@article{weiss2016survey,
  title={A survey of transfer learning},
  author={Weiss, Karl and Khoshgoftaar, Taghi M and Wang, DingDing},
  journal={Journal of Big data},
  volume={3},
  number={1},
  pages={9},
  year={2016},
  publisher={Springer}
}

@article{pan2009survey,
  title={A survey on transfer learning},
  author={Pan, Sinno Jialin and Yang, Qiang},
  journal={IEEE Transactions on knowledge and data engineering},
  volume={22},
  number={10},
  pages={1345--1359},
  year={2009},
  publisher={IEEE}
}

@inproceedings{tan2018survey,
  title={A survey on deep transfer learning},
  author={Tan, Chuanqi and Sun, Fuchun and Kong, Tao and Zhang, Wenchang and Yang, Chao and Liu, Chunfang},
  booktitle={Artificial Neural Networks and Machine Learning--ICANN 2018: 27th International Conference on Artificial Neural Networks, Rhodes, Greece, October 4-7, 2018, Proceedings, Part III 27},
  pages={270--279},
  year={2018},
  organization={Springer}
}

@inproceedings{mormont2018comparison,
  title={Comparison of deep transfer learning strategies for digital pathology},
  author={Mormont, Romain and Geurts, Pierre and Mar{\'e}e, Rapha{\"e}l},
  booktitle={Proceedings of the IEEE Conference on Computer Vision and Pattern Recognition Workshops},
  pages={2262--2271},
  year={2018}
}

@article{mazurowski2019deep,
  title={Deep learning in radiology: An overview of the concepts and a survey of the state of the art with focus on MRI},
  author={Mazurowski, Maciej A and Buda, Mateusz and Saha, Ashirbani and Bashir, Mustafa R},
  journal={Journal of magnetic resonance imaging},
  volume={49},
  number={4},
  pages={939--954},
  year={2019},
  publisher={Wiley Online Library}
}

@article{kraus2017automated,
  title={Automated analysis of high-content microscopy data with deep learning},
  author={Kraus, Oren Z and Grys, Ben T and Ba, Jimmy and Chong, Yolanda and Frey, Brendan J and Boone, Charles and Andrews, Brenda J},
  journal={Molecular systems biology},
  volume={13},
  number={4},
  pages={924},
  year={2017}
}

@inproceedings{sabatelli2018deep,
  title={Deep transfer learning for art classification problems},
  author={Sabatelli, Matthia and Kestemont, Mike and Daelemans, Walter and Geurts, Pierre},
  booktitle={Proceedings of the European Conference on Computer Vision (ECCV)},
  pages={0--0},
  year={2018}
}

@article{li2020deep,
  title={Deep facial expression recognition: A survey},
  author={Li, Shan and Deng, Weihong},
  journal={IEEE Transactions on Affective Computing},
  year={2020},
  publisher={IEEE}
}

@inproceedings{ng2015deep,
  title={Deep learning for emotion recognition on small datasets using transfer learning},
  author={Ng, Hong-Wei and Nguyen, Viet Dung and Vonikakis, Vassilios and Winkler, Stefan},
  booktitle={Proceedings of the 2015 ACM on international conference on multimodal interaction},
  pages={443--449},
  year={2015}
}

@article{minaee2020image,
  title = {Image Segmentation Using Deep Learning: A Survey},
   author = {Minaee, Shervin and Boykov, Yuri and Porikli, Fatih and Plaza, Antonio and Kehtarnavaz, Nasser and Terzopoulos, Demetri},
   journal = {IEEE Transactions on Pattern Analysis and Machine Intelligence},
   volume = {44},
   number = {7},
   pages = {3523-3542},
   ISSN = {1939-3539},
   DOI = {10.1109/TPAMI.2021.3059968},
   year = {2022},
   type = {Journal Article}
}

@article{ghosh2019understanding,
  title={Understanding deep learning techniques for image segmentation},
  author={Ghosh, Swarnendu and Das, Nibaran and Das, Ishita and Maulik, Ujjwal},
  journal={ACM Computing Surveys (CSUR)},
  volume={52},
  number={4},
  pages={1--35},
  year={2019},
  publisher={ACM New York, NY, USA}
}

@inproceedings{zoph2020rethinking,
  title = {Rethinking Pre-training and Self-training},
   author = {Zoph, Barret and Ghiasi, Golnaz and Lin, Tsung-Yi and Cui, Yin and Liu, Hanxiao and Cubuk, Ekin Dogus and Le, Quoc},
   booktitle = {Advances in Neural Information Processing Systems},
   publisher = {Curran Associates, Inc.},
   volume = {33},
   pages = {3833–3845},
   year = {2020},
   type = {Conference Proceedings}
}

@inproceedings{chen2018encoder,
  title={Encoder-decoder with atrous separable convolution for semantic image segmentation},
  author={Chen, Liang-Chieh and Zhu, Yukun and Papandreou, George and Schroff, Florian and Adam, Hartwig},
  booktitle={Proceedings of the European conference on computer vision (ECCV)},
  pages={801--818},
  year={2018}
}

@article{zhang2017transfer,
  title = {Recent Advances in Transfer Learning for Cross-Dataset Visual Recognition: A Problem-Oriented Perspective},
   author = {Zhang, Jing and Li, Wanqing and Ogunbona, Philip and Xu, Dong},
   journal = {ACM Comput. Surv.},
   volume = {52},
   number = {1},
   pages = {7:1–7:38},
   ISSN = {0360-0300},
   DOI = {10.1145/3291124},
   year = {2019},
   type = {Journal Article}
}

@article{wang2018deep1,
  title={Deep visual domain adaptation: A survey},
  author={Wang, Mei and Deng, Weihong},
  journal={Neurocomputing},
  volume={312},
  pages={135--153},
  year={2018},
  publisher={Elsevier}
}

@article{kaya2019deep,
  title={Deep metric learning: A survey},
  author={Kaya, Mahmut and Bilge, Hasan {\c{S}}akir},
  journal={Symmetry},
  volume={11},
  number={9},
  pages={1066},
  year={2019},
  publisher={Multidisciplinary Digital Publishing Institute}
}

@book{goodfellow2016deep,
  title={Deep learning},
  author={Goodfellow, Ian and Bengio, Yoshua and Courville, Aaron and Bengio, Yoshua},
  volume={1},
  year={2016},
  publisher={MIT press Cambridge}
}

@article{hinton2015distilling,
  title={Distilling the knowledge in a neural network},
  author={Hinton, Geoffrey and Vinyals, Oriol and Dean, Jeff},
  journal={arXiv preprint arXiv:1503.02531},
  year={2015}
}

@article{miller1995wordnet,
  title={WordNet: a lexical database for English},
  author={Miller, George A},
  journal={Communications of the ACM},
  volume={38},
  number={11},
  pages={39--41},
  year={1995},
  publisher={ACM New York, NY, USA}
}

@techreport{krizhevsky2009learning,
  title = {Learning Multiple Layers of Features from Tiny Images},
  author = {Krizhevsky, Alex},
  institution = {University of Toronto},
  year = {2009},
  number = {TR-2009},
  type = {Technical Report},
  url = {https://www.cs.toronto.edu/~kriz/learning-features-2009-TR.pdf}
}

@misc{pascal-voc-2007,
	author = "Everingham, M. and Van~Gool, L. and Williams, C. K. I. and Winn, J. and Zisserman, A.",
	title = "The {PASCAL} {V}isual {O}bject {C}lasses {C}hallenge 2007 {(VOC2007)} {R}esults",
	howpublished = "http://www.pascal-network.org /challenges/VOC/voc2007/workshop/index.html"}

@inproceedings{fei2004learning,
  title={Learning generative visual models from few training examples: An incremental bayesian approach tested on 101 object categories},
  author={Fei-Fei, Li and Fergus, Rob and Perona, Pietro},
  booktitle={2004 conference on computer vision and pattern recognition workshop},
  pages={178--178},
  year={2004},
  organization={IEEE}
}

@techreport{griffin2007caltech,
  title={Caltech-256 object category dataset},
  author={Griffin, Gregory and Holub, Alex and Perona, Pietro and others},
  year={2007},
  institution={Technical Report 7694, California Institute of Technology Pasadena}
}

@inproceedings{khosla2011novel,
  title={Novel dataset for fine-grained image categorization: Stanford dogs},
  author={Khosla, Aditya and Jayadevaprakash, Nityananda and Yao, Bangpeng and Li, Fei-Fei},
  booktitle={Proc. CVPR Workshop on Fine-Grained Visual Categorization (FGVC)},
  volume={2},
  number={1},
  year={2011}
}

@techreport{wah2011caltech,
  title = {The Caltech-UCSD Birds-200-2011 Dataset},
  author = {Wah, Catherine and Branson, Steve and Welinder, Peter and Perona, Pietro and Belongie, Serge},
  institution = {California Institute of Technology},
  year = {2011},
  number = {CNS-TR-2011-001},
  url = {http://www.vision.caltech.edu/datasets/cub_200_2011/}
}

@inproceedings{xiao2010sun,
  title={Sun database: Large-scale scene recognition from abbey to zoo},
  author={Xiao, Jianxiong and Hays, James and Ehinger, Krista A and Oliva, Aude and Torralba, Antonio},
  booktitle={2010 IEEE computer society conference on computer vision and pattern recognition},
  pages={3485--3492},
  year={2010},
  organization={IEEE}
}

@article{zhou2017places,
  title={Places: A 10 million image database for scene recognition},
  author={Zhou, Bolei and Lapedriza, Agata and Khosla, Aditya and Oliva, Aude and Torralba, Antonio},
  journal={IEEE transactions on pattern analysis and machine intelligence},
  volume={40},
  number={6},
  pages={1452--1464},
  year={2017},
  publisher={IEEE}
}

@inproceedings{quattoni2009recognizing,
  title={Recognizing indoor scenes},
  author={Quattoni, Ariadna and Torralba, Antonio},
  booktitle={2009 IEEE Conference on Computer Vision and Pattern Recognition},
  pages={413--420},
  year={2009},
  organization={IEEE}
}

@inproceedings{cimpoi2014describing,
  title={Describing textures in the wild},
  author={Cimpoi, Mircea and Maji, Subhransu and Kokkinos, Iasonas and Mohamed, Sammy and Vedaldi, Andrea},
  booktitle={Proceedings of the IEEE Conference on Computer Vision and Pattern Recognition},
  pages={3606--3613},
  year={2014}
}

@article{munder2006experimental,
  title={An experimental study on pedestrian classification},
  author={Munder, Stefan and Gavrila, Dariu M},
  journal={IEEE transactions on pattern analysis and machine intelligence},
  volume={28},
  number={11},
  pages={1863--1868},
  year={2006},
  publisher={IEEE}
}

@article{stallkamp2012man,
  title={Man vs. computer: Benchmarking machine learning algorithms for traffic sign recognition},
  author={Stallkamp, Johannes and Schlipsing, Marc and Salmen, Jan and Igel, Christian},
  journal={Neural networks},
  volume={32},
  pages={323--332},
  year={2012},
  publisher={Elsevier}
}

@article{lake2015human,
  title={Human-level concept learning through probabilistic program induction},
  author={Lake, Brenden M and Salakhutdinov, Ruslan and Tenenbaum, Joshua B},
  journal={Science},
  volume={350},
  number={6266},
  pages={1332--1338},
  year={2015},
  publisher={American Association for the Advancement of Science}
}

@inproceedings{netzer2011reading,
  title={Reading digits in natural images with unsupervised feature learning},
  author={Netzer, Yuval and Wang, Tao and Coates, Adam and Bissacco, Alessandro and Wu, Baolin and Ng, Andrew Y and others},
  booktitle={NIPS workshop on deep learning and unsupervised feature learning},
  volume={2011},
  number={2},
  pages={4},
  year={2011},
  organization={Granada}
}

@article{soomro2012ucf101,
  title={UCF101: A dataset of 101 human actions classes from videos in the wild},
  author={Soomro, Khurram and Zamir, Amir Roshan and Shah, Mubarak},
  journal={arXiv preprint arXiv:1212.0402},
  year={2012}
}

@inproceedings{rebuffi2017learning,
  title={Learning multiple visual domains with residual adapters},
  author={Rebuffi, Sylvestre-Alvise and Bilen, Hakan and Vedaldi, Andrea},
  booktitle={Advances in Neural Information Processing Systems},
  pages={506--516},
  year={2017}
}

@inproceedings{lin2014microsoft,
  title={Microsoft coco: Common objects in context},
  author={Lin, Tsung-Yi and Maire, Michael and Belongie, Serge and Hays, James and Perona, Pietro and Ramanan, Deva and Doll{\'a}r, Piotr and Zitnick, C Lawrence},
  booktitle={European conference on computer vision},
  pages={740--755},
  year={2014},
  organization={Springer}
}

@article{li2019delta,
  title = {Knowledge Distillation with Attention for Deep Transfer Learning of Convolutional Networks},
   author = {Li, Xingjian and Xiong, Haoyi and Chen, Zeyu and Huan, Jun and Liu, Ji and Xu, Cheng-Zhong and Dou, Dejing},
   journal = {ACM Trans. Knowl. Discov. Data},
   volume = {16},
   number = {3},
   pages = {42:1–42:20},
   ISSN = {1556-4681},
   DOI = {10.1145/3473912},
   year = {2021},
   type = {Journal Article}
}

@inproceedings{xuhong2018explicit,
  title={Explicit inductive bias for transfer learning with convolutional networks},
  author={Xuhong, LI and Grandvalet, Yves and Davoine, Franck},
  booktitle={International Conference on Machine Learning},
  pages={2825--2834},
  year={2018},
  organization={PMLR}
}

@article{li2020baseline,
  title={A baseline regularization scheme for transfer learning with convolutional neural networks},
  author={Li, Xuhong and Grandvalet, Yves and Davoine, Franck},
  journal={Pattern Recognition},
  volume={98},
  pages={107049},
  year={2020},
  publisher={Elsevier}
}

@inproceedings{wan2019towards,
  title={Towards Making Deep Transfer Learning Never Hurt},
  author={Wan, Ruosi and Xiong, Haoyi and Li, Xingjian and Zhu, Zhanxing and Huan, Jun},
  booktitle={2019 IEEE International Conference on Data Mining (ICDM)},
  pages={578--587},
  year={2019},
  organization={IEEE}
}

@inproceedings{li2020rethinking,
  title = {Rethinking the Hyperparameters for Fine-tuning},
   author = {Li, Hao and Chaudhari, Pratik and Yang, Hao and Lam, Michael and Ravichandran, Avinash and Bhotika, Rahul and Soatto, Stefano},
   booktitle = {International Conference on Learning Representations},
   year = {2019},
   type = {Conference Proceedings}
}

@inproceedings{kolesnikov2020big,
  title={Big transfer (bit): General visual representation learning},
  author={Kolesnikov, Alexander and Beyer, Lucas and Zhai, Xiaohua and Puigcerver, Joan and Yung, Jessica and Gelly, Sylvain and Houlsby, Neil},
  booktitle={Computer Vision--ECCV 2020: 16th European Conference, Glasgow, UK, August 23--28, 2020, Proceedings, Part V 16},
  pages={491--507},
  year={2020},
  organization={Springer}
}

@article{jeon2020sample,
  title={Sample-based Regularization: A Transfer Learning Strategy Toward Better Generalization},
  author={Jeon, Yunho and Choi, Yongseok and Park, Jaesun and Yi, Subin and Cho, Dongyeon and Kim, Jiwon},
  journal={arXiv preprint arXiv:2007.05181},
  year={2020}
}

@article{strubell2019energy,
  title = {Energy and Policy Considerations for Modern Deep Learning Research},
   author = {Strubell, Emma and Ganesh, Ananya and McCallum, Andrew},
   journal = {Proceedings of the AAAI Conference on Artificial Intelligence},
   volume = {34},
   number = {09},
   pages = {13693-13696},
   ISSN = {2374-3468},
   DOI = {10.1609/aaai.v34i09.7123},
   year = {2020},
   type = {Journal Article}
}

@inproceedings{bottou2008tradeoffs,
  title={The tradeoffs of large scale learning},
  author={Bottou, L{\'e}on and Bousquet, Olivier},
  booktitle={Advances in neural information processing systems},
  pages={161--168},
  year={2008}
}

@inproceedings{ioffe2015batch,
  title = {Batch normalization: accelerating deep network training by reducing internal covariate shift},
   author = {Ioffe, Sergey and Szegedy, Christian},
   booktitle = {Proceedings of the 32nd International Conference on International Conference on Machine Learning - Volume 37},
   series = {ICML'15},
   address = {Lille, France},
   publisher = {JMLR.org},
   pages = {448–456},
   year = {2015},
   type = {Conference Proceedings}
}

@inproceedings{tommasi2014testbed,
  title={A testbed for cross-dataset analysis},
  author={Tommasi, Tatiana and Tuytelaars, Tinne},
  booktitle={European Conference on Computer Vision},
  pages={18--31},
  year={2014},
  organization={Springer}
}

@inproceedings{saenko2010adapting,
  title={Adapting visual category models to new domains},
  author={Saenko, Kate and Kulis, Brian and Fritz, Mario and Darrell, Trevor},
  booktitle={European conference on computer vision},
  pages={213--226},
  year={2010},
  organization={Springer}
}

@inproceedings{gong2012geodesic,
  title={Geodesic flow kernel for unsupervised domain adaptation},
  author={Gong, Boqing and Shi, Yuan and Sha, Fei and Grauman, Kristen},
  booktitle={2012 IEEE Conference on Computer Vision and Pattern Recognition},
  pages={2066--2073},
  year={2012},
  organization={IEEE}
}

@inproceedings{chen2019catastrophic,
  title = {Catastrophic Forgetting Meets Negative Transfer: Batch Spectral Shrinkage for Safe Transfer Learning},
  author = {Chen, Xinyang and Wang, Sinan and Fu, Bo and Long, Mingsheng and Wang, Jianmin},
  booktitle = {Advances in Neural Information Processing Systems},
  volume = {32},
  year = {2019},
  url = {https://papers.nips.cc/paper_files/paper/2019/file/8466_catastrophic_forgetting_meets_negative_transfer_batch_spectral_shrinkage_for_safe_transfer_learning.pdf}
}

@inproceedings{liu2019towards,
  title = {Transferable Adversarial Training: A General Approach to Adapting Deep Classifiers},
   author = {Liu, Hong and Long, Mingsheng and Wang, Jianmin and Jordan, Michael},
   booktitle = {Proceedings of the 36th International Conference on Machine Learning},
   editor = {Chaudhuri, Kamalika and Salakhutdinov, Ruslan},
   series = {Proceedings of Machine Learning Research},
   publisher = {PMLR},
   volume = {97},
   pages = {4013–4022},
   year = {2019},
   type = {Conference Proceedings}
}

@article{you2020co,
  title={Co-tuning for transfer learning},
  author={You, Kaichao and Kou, Zhi and Long, Mingsheng and Wang, Jianmin},
  journal={Advances in Neural Information Processing Systems},
  volume={33},
  year={2020}
}

@article{kou2020stochastic,
  title={Stochastic Normalization},
  author={Kou, Zhi and You, Kaichao and Long, Mingsheng and Wang, Jianmin},
  journal={Advances in Neural Information Processing Systems},
  volume={33},
  year={2020}
}

@inproceedings{ioffe2017batch,
  title = {Batch Renormalization: Towards Reducing Minibatch Dependence in Batch-Normalized Models},
   author = {Ioffe, Sergey},
   booktitle = {Proceedings of the 31st International Conference on Neural Information Processing Systems},
   series = {NIPS'17},
   address = {Red Hook, NY, USA},
   publisher = {Curran Associates Inc.},
   pages = {1942–1950},
   ISBN = {978-1-5108-6096-4},
   year = {2017},
   type = {Conference Proceedings}
}

@inproceedings{wang2019characterizing,
  title={Characterizing and avoiding negative transfer},
  author={Wang, Zirui and Dai, Zihang and P{\'o}czos, Barnab{\'a}s and Carbonell, Jaime},
  booktitle={Proceedings of the IEEE/CVF Conference on Computer Vision and Pattern Recognition},
  pages={11293--11302},
  year={2019}
}

@inproceedings{wang2019transferable,
  title = {Transferable Normalization: Towards Improving Transferability of Deep Neural Networks},
  author = {Wang, Ximei and Jin, Ying and Long, Mingsheng and Wang, Jianmin and Jordan, Michael I},
  booktitle = {Advances in Neural Information Processing Systems},
  volume = {32},
  year = {2019},
  url = {https://papers.nips.cc/paper_files/paper/2019/file/8c1218c9fefb5f3a5d3e0f3e6b2f9f3e-Paper.pdf}
}

@inproceedings{maria2017autodial,
  title={Autodial: Automatic domain alignment layers},
  author={Maria Carlucci, Fabio and Porzi, Lorenzo and Caputo, Barbara and Ricci, Elisa and Rota Bulo, Samuel},
  booktitle={Proceedings of the IEEE International Conference on Computer Vision},
  pages={5067--5075},
  year={2017}
}

@inproceedings{xie2020self,
  title={Self-training with noisy student improves imagenet classification},
  author={Xie, Qizhe and Luong, Minh-Thang and Hovy, Eduard and Le, Quoc V},
  booktitle={Proceedings of the IEEE/CVF Conference on Computer Vision and Pattern Recognition},
  pages={10687--10698},
  year={2020}
}

@article{yalniz2019billion,
  title={Billion-scale semi-supervised learning for image classification},
  author={Yalniz, I Zeki and J{\'e}gou, Herv{\'e} and Chen, Kan and Paluri, Manohar and Mahajan, Dhruv},
  journal={arXiv preprint arXiv:1905.00546},
  year={2019}
}

@inproceedings{neyshabur2020being,
  title = {What is being transferred in transfer learning?},
   author = {Neyshabur, Behnam and Sedghi, Hanie and Zhang, Chiyuan},
   booktitle = {Proceedings of the 34th International Conference on Neural Information Processing Systems},
   series = {NIPS '20},
   address = {Red Hook, NY, USA},
   publisher = {Curran Associates Inc.},
   pages = {512–523},
   ISBN = {978-1-7138-2954-6},
   year = {2020},
   type = {Conference Proceedings}
}

@inproceedings{zhu2020learning,
  title={Learning to transfer learn: Reinforcement learning-based selection for adaptive transfer learning},
  author={Zhu, Linchao and Ar{\i}k, Sercan {\"O} and Yang, Yi and Pfister, Tomas},
  booktitle={European Conference on Computer Vision},
  pages={342--358},
  year={2020},
  organization={Springer}
}

@inproceedings{yoon2020data,
  title={Data valuation using reinforcement learning},
  author={Yoon, Jinsung and Arik, Sercan and Pfister, Tomas},
  booktitle={International Conference on Machine Learning},
  pages={10842--10851},
  year={2020},
  organization={PMLR}
}

@inproceedings{plested2021rethinking,
  title={Rethinking Binary Hyperparameters for Deep Transfer Learning},
  author={Plested, Jo and Shen, Xuyang and Gedeon, Tom},
  booktitle={International Conference on Neural Information Processing},
  pages={463--475},
  year={2021},
  organization={Springer}
}

@inproceedings{ribani2019survey,
  title={A survey of transfer learning for convolutional neural networks},
  author={Ribani, Ricardo and Marengoni, Mauricio},
  booktitle={2019 32nd SIBGRAPI Conference on Graphics, Patterns and Images Tutorials (SIBGRAPI-T)},
  pages={47--57},
  year={2019},
  organization={IEEE}
}

@inproceedings{azizi2021big,
  title={Big self-supervised models advance medical image classification},
  author={Azizi, Shekoofeh and Mustafa, Basil and Ryan, Fiona and Beaver, Zachary and Freyberg, Jan and Deaton, Jonathan and Loh, Aaron and Karthikesalingam, Alan and Kornblith, Simon and Chen, Ting and others},
  booktitle={Proceedings of the IEEE/CVF international conference on computer vision},
  pages={3478--3488},
  year={2021}
}

@inproceedings{heker2020joint,
  title = {Hierarchical Fine-Tuning for joint Liver Lesion Segmentation and Lesion Classification in CT},
   author = {Heker, Michal and Ben-Cohen, Avi and Greenspan, Hayit},
   booktitle = {2019 41st Annual International Conference of the IEEE Engineering in Medicine and Biology Society (EMBC)},
   pages = {895-898},
   DOI = {10.1109/EMBC.2019.8857127},
   year = {2019},
   type = {Conference Proceedings}
}

@inbook{raghu2019transfusion,
  title = {Transfusion: Understanding Transfer Learning for Medical Imaging},
   author = {Raghu, Maithra and Zhang, Chiyuan and Kleinberg, Jon and Bengio, Samy},
   booktitle = {Proceedings of the 33rd International Conference on Neural Information Processing Systems},
   publisher = {Curran Associates Inc.},
   address = {Red Hook, NY, USA},
   pages = {3347–3357},
   year = {2019},
   type = {Book Section}
}

@inproceedings{gonthier2020analysis,
  title = {An Analysis of the Transfer Learning of Convolutional Neural Networks for Artistic Images},
   author = {Gonthier, Nicolas and Gousseau, Yann and Ladjal, Saïd},
   booktitle = {Pattern Recognition. ICPR International Workshops and Challenges},
   editor = {Del Bimbo, Alberto and Cucchiara, Rita and Sclaroff, Stan and Farinella, Giovanni Maria and Mei, Tao and Bertini, Marco and Escalante, Hugo Jair and Vezzani, Roberto},
   address = {Cham},
   publisher = {Springer International Publishing},
   pages = {546-561},
   ISBN = {978-3-030-68796-0},
   DOI = {10.1007/978-3-030-68796-0_39},
   year = {2021},
   type = {Conference Proceedings}
}

@inproceedings{hendrycks2019benchmarking,
  title = {Benchmarking Neural Network Robustness to Common Corruptions and Perturbations},
   author = {Hendrycks, Dan and Dietterich, Thomas},
   booktitle = {International Conference on Learning Representations},
   year = {2018},
   type = {Conference Proceedings}
}

@techreport{maji13fine-grained,
   title         = {Fine-Grained Visual Classification of Aircraft},
   author        = {S. Maji and J. Kannala and E. Rahtu
                    and M. Blaschko and A. Vedaldi},
   year          = {2013},
   archivePrefix = {arXiv},
   institution   = {Toyota Technological Institute at Chicago},
   eprint        = {1306.5151},
   primaryClass  = "cs-cv",
}

@inproceedings{hendrycks2021natural,
  title={Natural adversarial examples},
  author={Hendrycks, Dan and Zhao, Kevin and Basart, Steven and Steinhardt, Jacob and Song, Dawn},
  booktitle={Proceedings of the IEEE/CVF Conference on Computer Vision and Pattern Recognition},
  pages={15262--15271},
  year={2021}
}

@inproceedings{bossard14,
  title = {Food-101 -- Mining Discriminative Components with Random Forests},
  author = {Bossard, Lukas and Guillaumin, Matthieu and Van Gool, Luc},
  booktitle = {European Conference on Computer Vision},
  year = {2014}
}

@inproceedings{berg2014birdsnap,
  title={Birdsnap: Large-scale fine-grained visual categorization of birds},
  author={Berg, Thomas and Liu, Jiongxin and Woo Lee, Seung and Alexander, Michelle L and Jacobs, David W and Belhumeur, Peter N},
  booktitle={Proceedings of the IEEE Conference on Computer Vision and Pattern Recognition},
  pages={2011--2018},
  year={2014}
}

@inproceedings{nilsback2008automated,
  title={Automated flower classification over a large number of classes},
  author={Nilsback, Maria-Elena and Zisserman, Andrew},
  booktitle={2008 Sixth Indian Conference on Computer Vision, Graphics \& Image Processing},
  pages={722--729},
  year={2008},
  organization={IEEE}
}

@inproceedings{parkhi2012cats,
  title={Cats and dogs},
  author={Parkhi, Omkar M and Vedaldi, Andrea and Zisserman, Andrew and Jawahar, CV},
  booktitle={2012 IEEE conference on computer vision and pattern recognition},
  pages={3498--3505},
  year={2012},
  organization={IEEE}
}

@inproceedings{KrauseStarkDengFei-Fei_3DRR2013,
  title = {3D Object Representations for Fine-Grained Categorization},
  booktitle = {4th International IEEE Workshop on  3D Representation and Recognition (3dRR-13)},
  year = {2013},
  address = {Sydney, Australia},
  author = {Jonathan Krause and Michael Stark and Jia Deng and Li Fei-Fei}
}

@inproceedings{he2016deep,
  title={Deep residual learning for image recognition},
  author={He, Kaiming and Zhang, Xiangyu and Ren, Shaoqing and Sun, Jian},
  booktitle={Proceedings of the IEEE conference on computer vision and pattern recognition},
  pages={770--778},
  year={2016}
}

@inproceedings{tran2019transferability,
  title={Transferability and hardness of supervised classification tasks},
  author={Tran, Anh T and Nguyen, Cuong V and Hassner, Tal},
  booktitle={Proceedings of the IEEE/CVF International Conference on Computer Vision},
  pages={1395--1405},
  year={2019}
}

@inproceedings{bao2019information,
  title={An information-theoretic approach to transferability in task transfer learning},
  author={Bao, Yajie and Li, Yang and Huang, Shao-Lun and Zhang, Lin and Zheng, Lizhong and Zamir, Amir and Guibas, Leonidas},
  booktitle={2019 IEEE International Conference on Image Processing (ICIP)},
  pages={2309--2313},
  year={2019},
  organization={IEEE}
}

@inproceedings{nguyen2020leep,
  title={Leep: A new measure to evaluate transferability of learned representations},
  author={Nguyen, Cuong and Hassner, Tal and Seeger, Matthias and Archambeau, Cedric},
  booktitle={International Conference on Machine Learning},
  pages={7294--7305},
  year={2020},
  organization={PMLR}
}

@inproceedings{wang2020multitune,
  title={MultiTune: Adaptive Integration of Multiple Fine-Tuning Models for Image Classification},
  author={Wang, Yu and Plested, Jo and Gedeon, Tom},
  booktitle={International Conference on Neural Information Processing},
  pages={488--496},
  year={2020},
  organization={Springer}
}

@inproceedings{van2018inaturalist,
  title={The inaturalist species classification and detection dataset},
  author={Van Horn, Grant and Mac Aodha, Oisin and Song, Yang and Cui, Yin and Sun, Chen and Shepard, Alex and Adam, Hartwig and Perona, Pietro and Belongie, Serge},
  booktitle={Proceedings of the IEEE conference on computer vision and pattern recognition},
  pages={8769--8778},
  year={2018}
}

@inproceedings{pathak2016context,
  title={Context encoders: Feature learning by inpainting},
  author={Pathak, Deepak and Krahenbuhl, Philipp and Donahue, Jeff and Darrell, Trevor and Efros, Alexei A},
  booktitle={Proceedings of the IEEE conference on computer vision and pattern recognition},
  pages={2536--2544},
  year={2016}
}

@inproceedings{ledig2017photo,
  title={Photo-realistic single image super-resolution using a generative adversarial network},
  author={Ledig, Christian and Theis, Lucas and Husz{\'a}r, Ferenc and Caballero, Jose and Cunningham, Andrew and Acosta, Alejandro and Aitken, Andrew and Tejani, Alykhan and Totz, Johannes and Wang, Zehan and others},
  booktitle={Proceedings of the IEEE conference on computer vision and pattern recognition},
  pages={4681--4690},
  year={2017}
}

@article{jing2020self,
  title={Self-supervised visual feature learning with deep neural networks: A survey},
  author={Jing, Longlong and Tian, Yingli},
  journal={IEEE transactions on pattern analysis and machine intelligence},
  year={2020},
  publisher={IEEE}
}

@article{radford2015unsupervised,
  title={Unsupervised representation learning with deep convolutional generative adversarial networks},
  author={Radford, Alec and Metz, Luke and Chintala, Soumith},
  journal={arXiv preprint arXiv:1511.06434},
  year={2015}
}

@article{goodfellow2014generative,
  title={Generative adversarial nets},
  author={Goodfellow, Ian and Pouget-Abadie, Jean and Mirza, Mehdi and Xu, Bing and Warde-Farley, David and Ozair, Sherjil and Courville, Aaron and Bengio, Yoshua},
  journal={Advances in neural information processing systems},
  volume={27},
  year={2014}
}

@inproceedings{shao2019objects365,
  title={Objects365: A large-scale, high-quality dataset for object detection},
  author={Shao, Shuai and Li, Zeming and Zhang, Tianyuan and Peng, Chao and Yu, Gang and Zhang, Xiangyu and Li, Jing and Sun, Jian},
  booktitle={Proceedings of the IEEE/CVF international conference on computer vision},
  pages={8430--8439},
  year={2019}
}

@article{kuznetsova2020open,
  title={The open images dataset v4},
  author={Kuznetsova, Alina and Rom, Hassan and Alldrin, Neil and Uijlings, Jasper and Krasin, Ivan and Pont-Tuset, Jordi and Kamali, Shahab and Popov, Stefan and Malloci, Matteo and Kolesnikov, Alexander and others},
  journal={International Journal of Computer Vision},
  volume={128},
  number={7},
  pages={1956--1981},
  year={2020},
  publisher={Springer}
}

@inproceedings{cai2022bigdetection,
  title={BigDetection: A Large-scale Benchmark for Improved Object Detector Pre-training},
  author={Cai, Likun and Zhang, Zhi and Zhu, Yi and Zhang, Li and Li, Mu and Xue, Xiangyang},
  booktitle={Proceedings of the IEEE/CVF Conference on Computer Vision and Pattern Recognition},
  pages={4777--4787},
  year={2022}
}

@article{liu2022improved,
  title={Improved Fine-Tuning by Better Leveraging Pre-Training Data},
  author={Liu, Ziquan and Xu, Yi and Xu, Yuanhong and Qian, Qi and Li, Hao and Ji, Xiangyang and Chan, Antoni and Jin, Rong},
  journal={Advances in Neural Information Processing Systems},
  volume={35},
  pages={32568--32581},
  year={2022}
}

@article{shen2022amf,
  title={AMF: Adaptable Weighting Fusion with Multiple Fine-tuning for Image Classification},
  author={Shen, Xuyang and Plested, Jo and Caldwell, Sabrina and Zhong, Yiran and Gedeon, Tom},
  journal={arXiv preprint arXiv:2207.12944},
  year={2022}
}

@article{fu2022ferm,
  title={FERM: A FEature-space Representation Measure for Improved Model Evaluation},
  author={Fu, Yeu Shin and Ge, Wenbo and Plested, Jo},
  journal={work},
  volume={10},
  pages={11},
  year={2022}
}

@article{gao2022large,
  title={Large-scale unsupervised semantic segmentation},
  author={Gao, Shanghua and Li, Zhong-Yu and Yang, Ming-Hsuan and Cheng, Ming-Ming and Han, Junwei and Torr, Philip},
  journal={IEEE Transactions on Pattern Analysis and Machine Intelligence},
  year={2022},
  publisher={IEEE}
}

@article{mo2022review,
  title={Review the state-of-the-art technologies of semantic segmentation based on deep learning},
  author={Mo, Yujian and Wu, Yan and Yang, Xinneng and Liu, Feilin and Liao, Yujun},
  journal={Neurocomputing},
  volume={493},
  pages={626--646},
  year={2022},
  publisher={Elsevier}
}

@inproceedings{dosovitskiy2020image,
  title = {An Image is Worth 16x16 Words: Transformers for Image Recognition at Scale},
   author = {Dosovitskiy, Alexey and Beyer, Lucas and Kolesnikov, Alexander and Weissenborn, Dirk and Zhai, Xiaohua and Unterthiner, Thomas and Dehghani, Mostafa and Minderer, Matthias and Heigold, Georg and Gelly, Sylvain and Uszkoreit, Jakob and Houlsby, Neil},
   booktitle = {International Conference on Learning Representations},
   year = {2020},
   type = {Conference Proceedings}
}

@article{yu2022coca,
  title = {CoCa: Contrastive Captioners are Image-Text Foundation Models},
   author = {Yu, Jiahui and Wang, Zirui and Vasudevan, Vijay and Yeung, Legg and Seyedhosseini, Mojtaba and Wu, Yonghui},
   journal = {Transactions on Machine Learning Research},
   ISSN = {2835-8856},
   year = {2022},
   type = {Journal Article}
}

@inproceedings{chen2023symbolic,
  title = {Symbolic discovery of optimization algorithms},
   author = {Chen, Xiangning and Liang, Chen and Huang, Da and Real, Esteban and Wang, Kaiyuan and Pham, Hieu and Dong, Xuanyi and Luong, Thang and Hsieh, Cho-Jui and Lu, Yifeng and Le, Quoc V.},
   booktitle = {Proceedings of the 37th International Conference on Neural Information Processing Systems},
   series = {NIPS '23},
   address = {Red Hook, NY, USA},
   publisher = {Curran Associates Inc.},
   pages = {49205–49233},
   year = {2023},
   type = {Conference Proceedings}
}

@inproceedings{ridnik2023ml,
  title={Ml-decoder: Scalable and versatile classification head},
  author={Ridnik, Tal and Sharir, Gilad and Ben-Cohen, Avi and Ben-Baruch, Emanuel and Noy, Asaf},
  booktitle={Proceedings of the IEEE/CVF Winter Conference on Applications of Computer Vision},
  pages={32--41},
  year={2023}
}

@article{liu2023learn,
  title={Learn from each other to Classify better: Cross-layer mutual attention learning for fine-grained visual classification},
  author={Liu, Dichao and Zhao, Longjiao and Wang, Yu and Kato, Jien},
  journal={Pattern Recognition},
  volume={140},
  pages={109550},
  year={2023},
  publisher={Elsevier}
}

@article{zhu2021transfer,
  title={Transfer learning of graph neural networks with ego-graph information maximization},
  author={Zhu, Qi and Yang, Carl and Xu, Yidan and Wang, Haonan and Zhang, Chao and Han, Jiawei},
  journal={Advances in Neural Information Processing Systems},
  volume={34},
  pages={1766--1779},
  year={2021}
}

@inproceedings{shrivastava2017learning,
  title={Learning from simulated and unsupervised images through adversarial training},
  author={Shrivastava, Ashish and Pfister, Tomas and Tuzel, Oncel and Susskind, Joshua and Wang, Wenda and Webb, Russell},
  booktitle={Proceedings of the IEEE conference on computer vision and pattern recognition},
  pages={2107--2116},
  year={2017}
}

@inproceedings{tzeng2017adversarial,
  title={Adversarial discriminative domain adaptation},
  author={Tzeng, Eric and Hoffman, Judy and Saenko, Kate and Darrell, Trevor},
  booktitle={Proceedings of the IEEE conference on computer vision and pattern recognition},
  pages={7167--7176},
  year={2017}
}

@article{li2019analysis,
  title={An analysis of pre-training on object detection},
  author={Li, Hengduo and Singh, Bharat and Najibi, Mahyar and Wu, Zuxuan and Davis, Larry S},
  journal={International Conference on Computer Vision Workshop 2021},
  year={2019}
}

@article{liu2016coupled,
  title={Coupled generative adversarial networks},
  author={Liu, Ming-Yu and Tuzel, Oncel},
  journal={Advances in neural information processing systems},
  volume={29},
  year={2016}
}

@inproceedings{taigman2017unsupervised,
  title={Unsupervised Cross-Domain Image Generation},
  author={Taigman, Yaniv and Polyak, Adam and Wolf, Lior},
  booktitle={International Conference on Learning Representations},
  year={2017}
}

@article{attia2020realistic,
  title={Realistic hair simulator for skin lesion images: A novel benchemarking tool},
  author={Attia, Mohamed and Hossny, Mohammed and Zhou, Hailing and Nahavandi, Saeid and Asadi, Hamed and Yazdabadi, Anousha},
  journal={Artificial Intelligence in Medicine},
  volume={108},
  pages={101933},
  year={2020},
  publisher={Elsevier}
}

@inproceedings{puigcerverscalable,
  title={Scalable Transfer Learning with Expert Models},
  author={Puigcerver, Joan and Ruiz, Carlos Riquelme and Mustafa, Basil and Renggli, Cedric and Pinto, Andr{\'e} Susano and Gelly, Sylvain and Keysers, Daniel and Houlsby, Neil},
  booktitle={International Conference on Learning Representations}, 
  year={2020}
}

@article{iman2023review,
  title={A review of deep transfer learning and recent advancements},
  author={Iman, Mohammadreza and Arabnia, Hamid Reza and Rasheed, Khaled},
  journal={Technologies},
  volume={11},
  number={2},
  pages={40},
  year={2023},
  publisher={MDPI}
}

@article{yu2022survey,
  title={A survey on deep transfer learning and beyond},
  author={Yu, Fuchao and Xiu, Xianchao and Li, Yunhui},
  journal={Mathematics},
  volume={10},
  number={19},
  pages={3619},
  year={2022},
  publisher={MDPI}
}

@article{panda2024transfer,
  title={Transfer Learning Applied to Computer Vision Problems: Survey on Current Progress, Limitations, and Opportunities},
  author={Panda, Aaryan and Panigrahi, Damodar and Mitra, Shaswata and Mittal, Sudip and Rahimi, Shahram},
  journal={arXiv preprint arXiv:2409.07736},
  year={2024}
}

@inproceedings{tzeng2014deep,
  title = {Adversarial Discriminative Domain Adaptation},
   author = {Tzeng, Eric and Hoffman, Judy and Saenko, Kate and Darrell, Trevor},
   booktitle = {2017 IEEE Conference on Computer Vision and Pattern Recognition (CVPR)},
   pages = {2962-2971},
   DOI = {10.1109/CVPR.2017.316},
   year = {2017},
   type = {Conference Proceedings}
}

@article{borgwardt2006integrating,
  title={Integrating structured biological data by kernel maximum mean discrepancy},
  author={Borgwardt, Karsten M and Gretton, Arthur and Rasch, Malte J and Kriegel, Hans-Peter and Sch{\"o}lkopf, Bernhard and Smola, Alex J},
  journal={Bioinformatics},
  volume={22},
  number={14},
  pages={e49--e57},
  year={2006},
  publisher={Oxford University Press}
}

@article{zhu2020deep,
  title={Deep subdomain adaptation network for image classification},
  author={Zhu, Yongchun and Zhuang, Fuzhen and Wang, Jindong and Ke, Guolin and Chen, Jingwu and Bian, Jiang and Xiong, Hui and He, Qing},
  journal={IEEE transactions on neural networks and learning systems},
  volume={32},
  number={4},
  pages={1713--1722},
  year={2020},
  publisher={IEEE}
}

@inproceedings{long2015learning,
  title={Learning transferable features with deep adaptation networks},
  author={Long, Mingsheng and Cao, Yue and Wang, Jianmin and Jordan, Michael},
  booktitle={International conference on machine learning},
  pages={97--105},
  year={2015},
  organization={PMLR}
}

@article{zhu2019multi,
  title={Multi-representation adaptation network for cross-domain image classification},
  author={Zhu, Yongchun and Zhuang, Fuzhen and Wang, Jindong and Chen, Jingwu and Shi, Zhiping and Wu, Wenjuan and He, Qing},
  journal={Neural Networks},
  volume={119},
  pages={214--221},
  year={2019},
  publisher={Elsevier}
}

@inproceedings{radford2021learning,
  title={Learning transferable visual models from natural language supervision},
  author={Radford, Alec and Kim, Jong Wook and Hallacy, Chris and Ramesh, Aditya and Goh, Gabriel and Agarwal, Sandhini and Sastry, Girish and Askell, Amanda and Mishkin, Pamela and Clark, Jack and others},
  booktitle={International conference on machine learning},
  pages={8748--8763},
  year={2021},
  organization={PMLR}
}

@inproceedings{gamrian2019transfer,
  title={Transfer learning for related reinforcement learning tasks via image-to-image translation},
  author={Gamrian, Shani and Goldberg, Yoav},
  booktitle={International conference on machine learning},
  pages={2063--2072},
  year={2019},
  organization={PMLR}
}

@article{el2021large,
  title={Are large-scale datasets necessary for self-supervised pre-training?},
  author={El-Nouby, Alaaeldin and Izacard, Gautier and Touvron, Hugo and Laptev, Ivan and Jegou, Herv'e and Grave, Edouard},
  journal={arXiv preprint arXiv:2112.10740},
  year={2021}
}

@article{mauricio2023comparing,
  title={Comparing vision transformers and convolutional neural networks for image classification: A literature review},
  author={Maur{\'\i}cio, Jos{\'e} and Domingues, In{\^e}s and Bernardino, Jorge},
  journal={Applied Sciences},
  volume={13},
  number={9},
  pages={5521},
  year={2023},
  publisher={MDPI}
}

@inproceedings{zhou2021convnets,
  title={Convnets vs. transformers: Whose visual representations are more transferable?},
  author={Zhou, Hong-Yu and Lu, Chixiang and Yang, Sibei and Yu, Yizhou},
  booktitle={Proceedings of the IEEE/CVF International Conference on Computer Vision},
  pages={2230--2238},
  year={2021}
}

@inproceedings{cherti2023reproducible,
  title={Reproducible scaling laws for contrastive language-image learning},
  author={Cherti, Mehdi and Beaumont, Romain and Wightman, Ross and Wortsman, Mitchell and Ilharco, Gabriel and Gordon, Cade and Schuhmann, Christoph and Schmidt, Ludwig and Jitsev, Jenia},
  booktitle={Proceedings of the IEEE/CVF conference on computer vision and pattern recognition},
  pages={2818--2829},
  year={2023}
}

@inproceedings{he2022masked,
  title={Masked autoencoders are scalable vision learners},
  author={He, Kaiming and Chen, Xinlei and Xie, Saining and Li, Yanghao and Doll{\'a}r, Piotr and Girshick, Ross},
  booktitle={Proceedings of the IEEE/CVF conference on computer vision and pattern recognition},
  pages={16000--16009},
  year={2022}
}

@article{salehi2023clip,
  title={CLIP meets Model Zoo Experts: Pseudo-Supervision for Visual Enhancement},
  author={Salehi, Mohammadreza and Farajtabar, Mehrdad and Horton, Maxwell and Faghri, Fartash and Pouransari, Hadi and Vemulapalli, Raviteja and Tuzel, Oncel and Farhadi, Ali and Rastegari, Mohammad and Mehta, Sachin},
  journal={arXiv preprint arXiv:2310.14108},
  year={2023}
}

@inproceedings{dai2024adaptive,
  title={Adaptive Domain-Enhanced Transfer Learning for Welding Defect Classification},
  author={Dai, Dan and Mohan, Anand and Franciosa, Pasquale and Zhang, Tong and Chen, CL Philip and Ceglarek, Dariusz},
  booktitle={2024 IEEE International Conference on Systems, Man, and Cybernetics (SMC)},
  pages={3152--3158},
  year={2024},
  organization={IEEE}
}

@inproceedings{you2021logme,
  title={Logme: Practical assessment of pre-trained models for transfer learning},
  author={You, Kaichao and Liu, Yong and Wang, Jianmin and Long, Mingsheng},
  booktitle={International Conference on Machine Learning},
  pages={12133--12143},
  year={2021},
  organization={PMLR}
}

@article{you2022ranking,
  title={Ranking and tuning pre-trained models: A new paradigm for exploiting model hubs},
  author={You, Kaichao and Liu, Yong and Zhang, Ziyang and Wang, Jianmin and Jordan, Michael I and Long, Mingsheng},
  journal={Journal of Machine Learning Research},
  volume={23},
  number={209},
  pages={1--47},
  year={2022}
}

@article{nagae2022automatic,
  title={Automatic layer selection for transfer learning and quantitative evaluation of layer effectiveness},
  author={Nagae, Satsuki and Kanda, Daigo and Kawai, Shin and Nobuhara, Hajime},
  journal={Neurocomputing},
  volume={469},
  pages={151--162},
  year={2022},
  publisher={Elsevier}
}

@inproceedings{davila2023gradient,
  title={Gradient-based fine-tuning strategy for improved transfer learning on surgical images},
  author={Davila, Ana and Colan, Jacinto and Hasegawa, Yasuhisa},
  booktitle={2023 International Symposium on Micro-NanoMehatronics and Human Science (MHS)},
  pages={1--5},
  year={2023},
  organization={IEEE}
}

@inproceedings{leesurgical,
  title={Surgical Fine-Tuning Improves Adaptation to Distribution Shifts},
  author={Lee, Yoonho and Chen, Annie S and Tajwar, Fahim and Kumar, Ananya and Yao, Huaxiu and Liang, Percy and Finn, Chelsea},
year={2023},
  booktitle={The Eleventh International Conference on Learning Representations}
}

@inproceedings{wortsman2022model,
  title={Model soups: averaging weights of multiple fine-tuned models improves accuracy without increasing inference time},
  author={Wortsman, Mitchell and Ilharco, Gabriel and Gadre, Samir Ya and Roelofs, Rebecca and Gontijo-Lopes, Raphael and Morcos, Ari S and Namkoong, Hongseok and Farhadi, Ali and Carmon, Yair and Kornblith, Simon and others},
  booktitle={International conference on machine learning},
  pages={23965--23998},
  year={2022},
  organization={PMLR}
}

@article{rame2022diverse,
  title={Diverse weight averaging for out-of-distribution generalization},
  author={Rame, Alexandre and Kirchmeyer, Matthieu and Rahier, Thibaud and Rakotomamonjy, Alain and Gallinari, Patrick and Cord, Matthieu},
  journal={Advances in Neural Information Processing Systems},
  volume={35},
  pages={10821--10836},
  year={2022}
}

@article{sadrtdinov2023stay,
  title={To stay or not to stay in the pre-train basin: Insights on ensembling in transfer learning},
  author={Sadrtdinov, Ildus and Pozdeev, Dmitrii and Vetrov, Dmitry P and Lobacheva, Ekaterina},
  journal={Advances in Neural Information Processing Systems},
  volume={36},
  pages={15936--15964},
  year={2023}
}

@article{antonio2025efficient,
  title={Efficient adaptive ensembling for image classification},
  author={Antonio, Bruno and Moroni, Davide and Martinelli, Massimo},
  journal={Expert Systems},
  volume={42},
  number={1},
  pages={e13424},
  year={2025},
  publisher={Wiley Online Library}
}

@inproceedings{reed2022self,
  title={Self-supervised pretraining improves self-supervised pretraining},
  author={Reed, Colorado J and Yue, Xiangyu and Nrusimha, Ani and Ebrahimi, Sayna and Vijaykumar, Vivek and Mao, Richard and Li, Bo and Zhang, Shanghang and Guillory, Devin and Metzger, Sean and others},
  booktitle={Proceedings of the IEEE/CVF Winter Conference on Applications of Computer Vision},
  pages={2584--2594},
  year={2022}
}

@article{huang2023self,
  title={Self-supervised learning for medical image classification: a systematic review and implementation guidelines},
  author={Huang, Shih-Cheng and Pareek, Anuj and Jensen, Malte and Lungren, Matthew P and Yeung, Serena and Chaudhari, Akshay S},
  journal={NPJ Digital Medicine},
  volume={6},
  number={1},
  pages={74},
  year={2023},
  publisher={Nature Publishing Group UK London}
}

@article{xie2021towards,
  title={Towards effective deep transfer via attentive feature alignment},
  author={Xie, Zheng and Wen, Zhiquan and Wang, Yaowei and Wu, Qingyao and Tan, Mingkui},
  journal={Neural Networks},
  volume={138},
  pages={98--109},
  year={2021},
  publisher={Elsevier}
}

@inproceedings{fengrethinking,
  title={Rethinking Supervised Pre-Training for Better Downstream Transferring},
  author={Feng, Yutong and Jiang, Jianwen and Tang, Mingqian and Jin, Rong and Gao, Yue},
year={2021},
  booktitle={International Conference on Learning Representations}
}

@inproceedings{fan2021adversarially,
  title={Adversarially adaptive normalization for single domain generalization},
  author={Fan, Xinjie and Wang, Qifei and Ke, Junjie and Yang, Feng and Gong, Boqing and Zhou, Mingyuan},
  booktitle={Proceedings of the IEEE/CVF conference on Computer Vision and Pattern Recognition},
  pages={8208--8217},
  year={2021}
}

@inproceedings{liuncertainty,
  title={Uncertainty Modeling for Out-of-Distribution Generalization},
  author={Li, Xiaotong and Dai, Yongxing and Ge, Yixiao and Liu, Jun and Shan, Ying and DUAN, LINGYU},
year={2022},
  booktitle={International Conference on Learning Representations}
}

@article{nado2020evaluating,
  title={Evaluating prediction-time batch normalization for robustness under covariate shift},
  author={Nado, Zachary and Padhy, Shreyas and Sculley, D and D'Amour, Alexander and Lakshminarayanan, Balaji and Snoek, Jasper},
  journal={arXiv preprint arXiv:2006.10963},
  year={2020}
}

@article{huang2023normalization,
  title={Normalization techniques in training dnns: Methodology, analysis and application},
  author={Huang, Lei and Qin, Jie and Zhou, Yi and Zhu, Fan and Liu, Li and Shao, Ling},
  journal={IEEE transactions on pattern analysis and machine intelligence},
  volume={45},
  number={8},
  pages={10173--10196},
  year={2023},
  publisher={IEEE}
}

@inproceedings{you2024efficient,
  title={Efficient ConvBN Blocks for Transfer Learning and Beyond},
  author={You, Kaichao and Qin, Guo and Bao, Anchang and Cao, Meng and Huang, Ping and Shan, Jiulong and Long, Mingsheng},
  booktitle={International Conference on Learning Representations},
  year={2024}
}

@inproceedings{baobeit,
  title={BEiT: BERT Pre-Training of Image Transformers},
  author={Bao, Hangbo and Dong, Li and Piao, Songhao and Wei, Furu},
year={2021},
  booktitle={International Conference on Learning Representations}
}

@inproceedings{ramesh2021zero,
  title={Zero-shot text-to-image generation},
  author={Ramesh, Aditya and Pavlov, Mikhail and Goh, Gabriel and Gray, Scott and Voss, Chelsea and Radford, Alec and Chen, Mark and Sutskever, Ilya},
  booktitle={International conference on machine learning},
  pages={8821--8831},
  year={2021},
  organization={Pmlr}
}

@article{jaquier2025transfer,
  title={Transfer learning in robotics: An upcoming breakthrough? A review of promises and challenges},
  author={Jaquier, No{\'e}mie and Welle, Michael C and Gams, Andrej and Yao, Kunpeng and Fichera, Bernardo and Billard, Aude and Ude, Ale{\v{s}} and Asfour, Tamim and Kragic, Danica},
  journal={The International Journal of Robotics Research},
  volume={44},
  number={3},
  pages={465--485},
  year={2025},
  publisher={SAGE Publications Sage UK: London, England}
}

@article{kora2022transfer,
  title={Transfer learning techniques for medical image analysis: A review},
  author={Kora, Padmavathi and Ooi, Chui Ping and Faust, Oliver and Raghavendra, U and Gudigar, Anjan and Chan, Wai Yee and Meenakshi, K and Swaraja, K and Plawiak, Pawel and Acharya, U Rajendra},
  journal={Biocybernetics and biomedical engineering},
  volume={42},
  number={1},
  pages={79--107},
  year={2022},
  publisher={Elsevier}
}

@article{yu2025survey,
  title={A Survey of Foundation Models for Environmental Science},
  author={Yu, Runlong and Chen, Shengyu and Xie, Yiqun and Jia, Xiaowei},
  journal={arXiv preprint arXiv:2503.03142},
  year={2025}
}

@article{hossen2025transfer,
  title={Transfer learning in agriculture: a review},
  author={Hossen, Md Ismail and Awrangjeb, Mohammad and Pan, Shirui and Mamun, Abdullah Al},
  journal={Artificial Intelligence Review},
  volume={58},
  number={4},
  pages={97},
  year={2025},
  publisher={Springer}
}

@inproceedings{wang2021advsim,
  title={Advsim: Generating safety-critical scenarios for self-driving vehicles},
  author={Wang, Jingkang and Pun, Ava and Tu, James and Manivasagam, Sivabalan and Sadat, Abbas and Casas, Sergio and Ren, Mengye and Urtasun, Raquel},
  booktitle={Proceedings of the IEEE/CVF Conference on Computer Vision and Pattern Recognition},
  pages={9909--9918},
  year={2021}
}

@article{gabriel2020artificial,
  title={Artificial intelligence, values, and alignment},
  author={Gabriel, Iason},
  journal={Minds and machines},
  volume={30},
  number={3},
  pages={411--437},
  year={2020},
  publisher={Springer}
}

@article{tazi2022pyrocast,
  title={Pyrocast: a machine learning pipeline to forecast pyrocumulonimbus (pyrocb) clouds},
  author={Tazi, Kenza and Salas-Porras, Emiliano Diaz and Braude, Ashwin and Okoh, Daniel and Lamb, Kara D and Watson-Parris, Duncan and Harder, Paula and Meinert, Nis},
  journal={arXiv preprint arXiv:2211.13052},
  year={2022}
}

@inproceedings{kingmaauto,
  title={Auto-encoding variational Bayes},
  author={Kingma, Diederik P and Welling, Max},
year={2014},
  booktitle={International Conference on Learning Representations}
}

@article{gui2024survey,
  title={A survey on self-supervised learning: Algorithms, applications, and future trends},
  author={Gui, Jie and Chen, Tuo and Zhang, Jing and Cao, Qiong and Sun, Zhenan and Luo, Hao and Tao, Dacheng},
  journal={IEEE Transactions on Pattern Analysis and Machine Intelligence},
  year={2024},
  publisher={IEEE}
}

@article{kumar2024medical,
  title={Medical images classification using deep learning: a survey},
  author={Kumar, Rakesh and Kumbharkar, Pooja and Vanam, Sandeep and Sharma, Sanjeev},
  journal={Multimedia Tools and Applications},
  volume={83},
  number={7},
  pages={19683--19728},
  year={2024},
  publisher={Springer}
}

@article{olorunfemi2024advancements,
  title={Advancements in machine visions for fruit sorting and grading: A bibliometric analysis, systematic review, and future research directions},
  author={Olorunfemi, Benjamin Oluwamuyiwa and Nwulu, Nnamdi I and Adebo, Oluwafemi Ayodeji and Kavadias, Kosmas A},
  journal={Journal of Agriculture and Food Research},
  volume={16},
  pages={101154},
  year={2024},
  publisher={Elsevier}
}

@article{du2024datamap,
  title={DataMap: Dataset transferability map for medical image classification},
  author={Du, Xiangtong and Liu, Zhidong and Feng, Zunlei and Deng, Hai},
  journal={Pattern Recognition},
  volume={146},
  pages={110044},
  year={2024},
  publisher={Elsevier}
}

@article{bolya2021scalable,
  title={Scalable diverse model selection for accessible transfer learning},
  author={Bolya, Daniel and Mittapalli, Rohit and Hoffman, Judy},
  journal={Advances in Neural Information Processing Systems},
  volume={34},
  pages={19301--19312},
  year={2021}
}

@article{guo2024kite,
  title={KITE: A kernel-based improved transferability estimation method},
  author={Guo, Yunhui},
  journal={arXiv preprint arXiv:2405.01603},
  year={2024}
}

@article{tan2024transferability,
  title={Transferability-guided cross-domain cross-task transfer learning},
  author={Tan, Yang and Zhang, Enming and Li, Yang and Huang, Shao-Lun and Zhang, Xiao-Ping},
  journal={IEEE Transactions on Neural Networks and Learning Systems},
  year={2024},
  publisher={IEEE}
}

@inproceedings{tan2021otce,
  title={Otce: A transferability metric for cross-domain cross-task representations},
  author={Tan, Yang and Li, Yang and Huang, Shao-Lun},
  booktitle={Proceedings of the IEEE/CVF conference on computer vision and pattern recognition},
  pages={15779--15788},
  year={2021}
}

@article{juodelyte2024dataset,
  title={On dataset transferability in medical image classification},
  author={Juodelyte, Dovile and Ferrante, Enzo and Lu, Yucheng and Singh, Prabhant and Vanschoren, Joaquin and Cheplygina, Veronika},
  journal={arXiv preprint arXiv:2412.20172},
  year={2024}
}

@article{yu2024self,
  title={Self-supervised transformation learning for equivariant representations},
  author={Yu, Jaemyung and Choi, Jaehyun and Lee, DongJae and Hong, HyeongGwon and Kim, Junmo},
  journal={Advances in Neural Information Processing Systems},
  volume={37},
  pages={83068--83090},
  year={2024}
}

@article{wang2024understanding,
  title={Understanding the role of equivariance in self-supervised learning},
  author={Wang, Yifei and Hu, Kaiwen and Gupta, Sharut and Ye, Ziyu and Wang, Yisen and Jegelka, Stefanie},
  journal={Advances in Neural Information Processing Systems},
  volume={37},
  pages={127483--127510},
  year={2024}
}

@article{zhu2024awt,
  title={Awt: Transferring vision-language models via augmentation, weighting, and transportation},
  author={Zhu, Yuhan and Ji, Yuyang and Zhao, Zhiyu and Wu, Gangshan and Wang, Limin},
  journal={Advances in Neural Information Processing Systems},
  volume={37},
  pages={25561--25591},
  year={2024}
}

@article{shen2024expanding,
  title={Expanding sparse tuning for low memory usage},
  author={Shen, Shufan and Sun, Junshu and Ji, Xiangyang and Huang, Qingming and Wang, Shuhui},
  journal={Advances in Neural Information Processing Systems},
  volume={37},
  pages={76616--76642},
  year={2024}
}

@article{xin2024v,
  title={V-petl bench: A unified visual parameter-efficient transfer learning benchmark},
  author={Xin, Yi and Luo, Siqi and Liu, Xuyang and Zhou, Haodi and Cheng, Xinyu and Lee, Christina E and Du, Junlong and Wang, Haozhe and Chen, MingCai and Liu, Ting and others},
  journal={Advances in neural information processing systems},
  volume={37},
  pages={80522--80535},
  year={2024}
}

@article{yu2024visual,
  title={Visual tuning},
  author={Yu, Bruce XB and Chang, Jianlong and Wang, Haixin and Liu, Lingbo and Wang, Shijie and Wang, Zhiyu and Lin, Junfan and Xie, Lingxi and Li, Haojie and Lin, Zhouchen and others},
  journal={ACM Computing Surveys},
  volume={56},
  number={12},
  pages={1--38},
  year={2024},
  publisher={ACM New York, NY}
}

@inbook{RN28,
   author = {Torrey, Lisa and Shavlik, Jude},
   title = {Transfer learning},
   booktitle = {Handbook of research on machine learning applications and trends: algorithms, methods, and techniques},
   publisher = {IGI Global Scientific Publishing},
   pages = {242-264},
   year = {2010},
   type = {Book Section}
}

@article{RN13,
   author = {Zhuang, Fuzhen and Qi, Zhiyuan and Duan, Keyu and Xi, Dongbo and Zhu, Yongchun and Zhu, Hengshu and Xiong, Hui and He, Qing},
   title = {A comprehensive survey on transfer learning},
   journal = {Proceedings of the IEEE},
   volume = {109},
   number = {1},
   pages = {43-76},
   ISSN = {0018-9219},
   year = {2020},
   type = {Journal Article}
}

@article{RN103,
   author = {Rozinek, Ondřej and Mareš, Jan},
   title = {The duality of similarity and metric spaces},
   journal = {Applied Sciences},
   volume = {11},
   number = {4},
   pages = {1910},
   ISSN = {2076-3417},
   year = {2021},
   type = {Journal Article}
}

@article{RN104,
   author = {Ben-David, Shai and Blitzer, John and Crammer, Koby and Kulesza, Alex and Pereira, Fernando and Vaughan, Jennifer Wortman},
   title = {A theory of learning from different domains},
   journal = {Machine learning},
   volume = {79},
   number = {1},
   pages = {151-175},
   ISSN = {0885-6125},
   year = {2010},
   type = {Journal Article}
}

@article{RN26,
   author = {Stolte, Marieke and Kappenberg, Franziska and Rahnenführer, Jörg and Bommert, Andrea},
   title = {Methods for quantifying dataset similarity: a review, taxonomy and comparison},
   journal = {Statistic Surveys},
   volume = {18},
   pages = {163-298},
   ISSN = {1935-7516},
   year = {2024},
   type = {Journal Article}
}

@article{RN31,
   author = {Gupta, Surendra and Thakar, Urjita and Tokekar, Sanjiv},
   title = {A comprehensive survey on techniques for numerical similarity measurement},
   journal = {Expert Systems with Applications},
   pages = {127235},
   ISSN = {0957-4174},
   year = {2025},
   type = {Journal Article}
}

@inproceedings{szegedy2015going,
  title={Going deeper with convolutions},
  author={Szegedy, Christian and Liu, Wei and Jia, Yangqing and Sermanet, Pierre and Reed, Scott and Anguelov, Dragomir and Erhan, Dumitru and Vanhoucke, Vincent and Rabinovich, Andrew},
  booktitle={Proceedings of the IEEE conference on computer vision and pattern recognition},
  pages={1--9},
  year={2015}
}

@article{mai2024lessons,
  title={Lessons learned from a unifying empirical study of parameter-efficient transfer learning (petl) in visual recognition},
  author={Mai, Zheda and Zhang, Ping and Tu, Cheng-Hao and Chen, Hong-You and Zhang, Li and Chao, Wei-Lun},
  year={2024},
  journal={Open Review}
}

@inproceedings{RN107,
   author = {Long, Mingsheng and Cao, Yue and Wang, Jianmin and Jordan, Michael},
   title = {Learning transferable features with deep adaptation networks},
   booktitle = {International conference on machine learning},
   publisher = {PMLR},
   pages = {97-105},
   year = {2015},
   type = {Conference Proceedings}
}

@article{RN108,
   author = {Cheng, Cheng and Zhou, Beitong and Ma, Guijun and Wu, Dongrui and Yuan, Ye},
   title = {Wasserstein distance based deep adversarial transfer learning for intelligent fault diagnosis with unlabeled or insufficient labeled data},
   journal = {Neurocomputing},
   volume = {409},
   pages = {35-45},
   ISSN = {0925-2312},
   year = {2019},
   type = {Journal Article}
}

@inproceedings{RN109,
   author = {Zhang, Chi and Cai, Yujun and Lin, Guosheng and Shen, Chunhua},
   title = {DeepEMD: Few-shot image classification with differentiable earth mover's distance and structured classifiers},
   booktitle = {Proceedings of the IEEE/CVF conference on computer vision and pattern recognition},
   pages = {12203-12213},
   year = {2020},
   type = {Conference Proceedings}
}

@inproceedings{RN110,
   author = {Chen, Cheng and Dou, Qi and Chen, Hao and Qin, Jing and Heng, Pheng-Ann},
   title = {Synergistic image and feature adaptation: Towards cross-modality domain adaptation for medical image segmentation},
   booktitle = {Proceedings of the AAAI conference on artificial intelligence},
   volume = {33},
   pages = {865-872},
   ISBN = {2374-3468},
   year = {2019},
   type = {Conference Proceedings}
}

@inproceedings{komodakis2018unsupervised,
  title={Unsupervised representation learning by predicting image rotations},
  author={Komodakis, Nikos and Gidaris, Spyros},
  booktitle={International Conference on Learning Representations (ICLR)},
  year={2018}
}

@inproceedings{noroozi2016unsupervised,
  title={Unsupervised learning of visual representations by solving jigsaw puzzles},
  author={Noroozi, Mehdi and Favaro, Paolo},
  booktitle={European conference on computer vision},
  pages={69--84},
  year={2016},
  organization={Springer}
}

@inproceedings{doersch2015unsupervised,
  title={Unsupervised visual representation learning by context prediction},
  author={Doersch, Carl and Gupta, Abhinav and Efros, Alexei A},
  booktitle={Proceedings of the IEEE international conference on computer vision},
  pages={1422--1430},
  year={2015}
}

@article{gupta2023structuring,
  title={Structuring Representation Geometry with Rotationally Equivariant Contrastive Learning},
  author={Gupta, Sharut and Robinson, Joshua and Lim, Derek and Villar, Soledad and Jegelka, Stefanie},
  journal={CoRR},
  year={2023}
}

@article{lee2021improving,
  title={Improving transferability of representations via augmentation-aware self-supervision},
  author={Lee, Hankook and Lee, Kibok and Lee, Kimin and Lee, Honglak and Shin, Jinwoo},
  journal={Advances in Neural Information Processing Systems},
  volume={34},
  pages={17710--17722},
  year={2021}
}

@inproceedings{dehghani2023scaling,
  title={Scaling vision transformers to 22 billion parameters},
  author={Dehghani, Mostafa and Djolonga, Josip and Mustafa, Basil and Padlewski, Piotr and Heek, Jonathan and Gilmer, Justin and Steiner, Andreas Peter and Caron, Mathilde and Geirhos, Robert and Alabdulmohsin, Ibrahim and others},
  booktitle={International conference on machine learning},
  pages={7480--7512},
  year={2023},
  organization={PMLR}
}

@article{hu2022lora,
  title={Lora: Low-rank adaptation of large language models.},
  author={Hu, Edward J and Shen, Yelong and Wallis, Phillip and Allen-Zhu, Zeyuan and Li, Yuanzhi and Wang, Shean and Wang, Lu and Chen, Weizhu and others},
  journal={ICLR},
  volume={1},
  number={2},
  pages={3},
  year={2022}
}

@article{zaken2021bitfit,
  title={Bitfit: Simple parameter-efficient fine-tuning for transformer-based masked language-models},
  author={Zaken, Elad Ben and Ravfogel, Shauli and Goldberg, Yoav},
  journal={arXiv preprint arXiv:2106.10199},
  year={2021}
}

@inproceedings{xie2023difffit,
  title={Difffit: Unlocking transferability of large diffusion models via simple parameter-efficient fine-tuning},
  author={Xie, Enze and Yao, Lewei and Shi, Han and Liu, Zhili and Zhou, Daquan and Liu, Zhaoqiang and Li, Jiawei and Li, Zhenguo},
  booktitle={Proceedings of the IEEE/CVF International Conference on Computer Vision},
  pages={4230--4239},
  year={2023}
}

@inproceedings{basu2024strong,
  title={Strong baselines for parameter-efficient few-shot fine-tuning},
  author={Basu, Samyadeep and Hu, Shell and Massiceti, Daniela and Feizi, Soheil},
  booktitle={Proceedings of the AAAI Conference on Artificial Intelligence},
  volume={38},
  number={10},
  pages={11024--11031},
  year={2024}
}

@article{lialin2023scaling,
  title={Scaling down to scale up: A guide to parameter-efficient fine-tuning},
  author={Lialin, Vladislav and Deshpande, Vijeta and Rumshisky, Anna},
  journal={arXiv preprint arXiv:2303.15647},
  year={2023}
}

@inproceedings{jia2022visual,
  title={Visual prompt tuning},
  author={Jia, Menglin and Tang, Luming and Chen, Bor-Chun and Cardie, Claire and Belongie, Serge and Hariharan, Bharath and Lim, Ser-Nam},
  booktitle={European conference on computer vision},
  pages={709--727},
  year={2022},
  organization={Springer}
}

@article{ding2024model,
  title={Which model to transfer? a survey on transferability estimation},
  author={Ding, Yuhe and Jiang, Bo and Yu, Aijing and Zheng, Aihua and Liang, Jian},
  journal={arXiv preprint arXiv:2402.15231},
  year={2024}
}

@article{qi2022transferability,
  title={Transferability estimation based on principal gradient expectation},
  author={Qi, Huiyan and Cheng, Lechao and Chen, Jingjing and Yu, Yue and Song, Xue and Feng, Zunlei and Jiang, Yu-Gang},
  journal={arXiv preprint arXiv:2211.16299},
  year={2022}
}

@inproceedings{agostinelli2022stable,
  title={How stable are transferability metrics evaluations?},
  author={Agostinelli, Andrea and P{\'a}ndy, Michal and Uijlings, Jasper and Mensink, Thomas and Ferrari, Vittorio},
  booktitle={European Conference on Computer Vision},
  pages={303--321},
  year={2022},
  organization={Springer}
}

@inproceedings{chaves2023performance,
  title={The performance of transferability metrics does not translate to medical tasks},
  author={Chaves, Levy and Bissoto, Alceu and Valle, Eduardo and Avila, Sandra},
  booktitle={MICCAI Workshop on Domain Adaptation and Representation Transfer},
  pages={105--114},
  year={2023},
  organization={Springer}
}

@article{RN182,
   author = {So, Ronald WK and Chung, Albert CS},
   title = {A novel learning-based dissimilarity metric for rigid and non-rigid medical image registration by using Bhattacharyya Distances},
   journal = {Pattern Recognition},
   volume = {62},
   pages = {161-174},
   ISSN = {0031-3203},
   year = {2017},
   type = {Journal Article}
}

@incollection{biswas2023generative,
  title={Generative adversarial networks for data augmentation},
  author={Biswas, Angona and Md Abdullah Al, Nasim and Imran, Al and Sejuty, Anika Tabassum and Fairooz, Fabliha and Puppala, Sai and Talukder, Sajedul},
  booktitle={Data Driven Approaches on Medical Imaging},
  pages={159--177},
  year={2023},
  publisher={Springer}
}

@inproceedings{ribeiro2016should,
  title={" Why should i trust you?" Explaining the predictions of any classifier},
  author={Ribeiro, Marco Tulio and Singh, Sameer and Guestrin, Carlos},
  booktitle={Proceedings of the 22nd ACM SIGKDD international conference on knowledge discovery and data mining},
  pages={1135--1144},
  year={2016}
}

@article{han2024parameter,
  title={Parameter-efficient fine-tuning for large models: A comprehensive survey},
  author={Han, Zeyu and Gao, Chao and Liu, Jinyang and Zhang, Jeff and Zhang, Sai Qian},
  journal={arXiv preprint arXiv:2403.14608},
  year={2024}
}

@article{zhong2024convolution,
  title={Convolution meets lora: Parameter efficient finetuning for segment anything model},
  author={Zhong, Zihan and Tang, Zhiqiang and He, Tong and Fang, Haoyang and Yuan, Chun},
  journal={arXiv preprint arXiv:2401.17868},
  year={2024}
}

\end{document}